\documentclass[journal]{IEEEtran}
\usepackage{microtype}                 
\PassOptionsToPackage{warn}{textcomp}  
\usepackage{tikz}
\usepackage{comment}
\usepackage{amsmath,amssymb} 
\usepackage{color}
\usepackage{lineno,hyperref}
\usepackage{comment}
\usepackage{bbm}
\usepackage{amsmath, xparse}
\usepackage{here}
\usepackage{xcolor}
\usepackage{subfig}
\usepackage{bm}
\usepackage{multirow}
\usepackage{ctable} 
\usepackage{arydshln}
\usepackage{algorithm}
\usepackage{algorithmic}
\newcommand\norm[1]{\left\lVert#1\right\rVert}
\newcommand{\specialcell}[2][c]{%
  \begin{tabular}[#1]{@{}c@{}}#2\end{tabular}}

\ifCLASSINFOpdf
\else
\fi
\begin{document}
%
\title{Learning Disentangled Representations for Controllable Human Motion Prediction}
%
%
%

\author{Chunzhi Gu,
        Jun Yu,
        and~Chao~Zhang
\thanks{Manuscript received XX, xx; revised XX. (Corresponding author: C. Zhang.) }
\thanks{C. Gu and C. Zhang* are with the School of Engineering, University of Fukui, Fukui, Japan (e-mails: gu-cz@monju.fuis.u-fukui.ac.jp; zhang@u-fukui.ac.jp).}
\thanks{J. Yu is with Institute of Science and Technology, Niigata University, Japan (e-mail: yujun@ie.niigata-u.ac.jp).}
}

\maketitle

\begin{abstract}
Generative model-based motion prediction techniques have recently realized predicting controlled human motions, such as predicting multiple upper human body motions with similar lower-body motions. However, to achieve this, the state-of-the-art methods require either subsequently learning mapping functions to seek similar motions or training the model repetitively to enable control over the desired portion of body. In this paper, we propose a novel framework to learn disentangled representations for controllable human motion prediction. Our network involves a conditional variational auto-encoder (CVAE) architecture to model full-body human motion, and an extra CVAE path to learn only the corresponding partial-body (e.g., lower-body) motion. Specifically, the inductive bias imposed by the extra CVAE path encourages two latent variables in two paths to respectively govern separate representations for each partial-body motion. With a single training, our model is able to provide two types of controls for the generated human motions: (i) strictly controlling one portion of human body and (ii) adaptively controlling the other portion, by sampling from a pair of latent spaces. Additionally, we extend and adapt a sampling strategy to our trained model to diversify the controllable predictions. Our framework also potentially allows new forms of control by flexibly customizing the input for the extra CVAE path. Extensive experimental results and ablation studies demonstrate that our approach is capable of predicting state-of-the-art controllable human motions both qualitatively and quantitatively. 
\end{abstract}

\begin{figure*}
\centering
\includegraphics[width=0.92\linewidth]{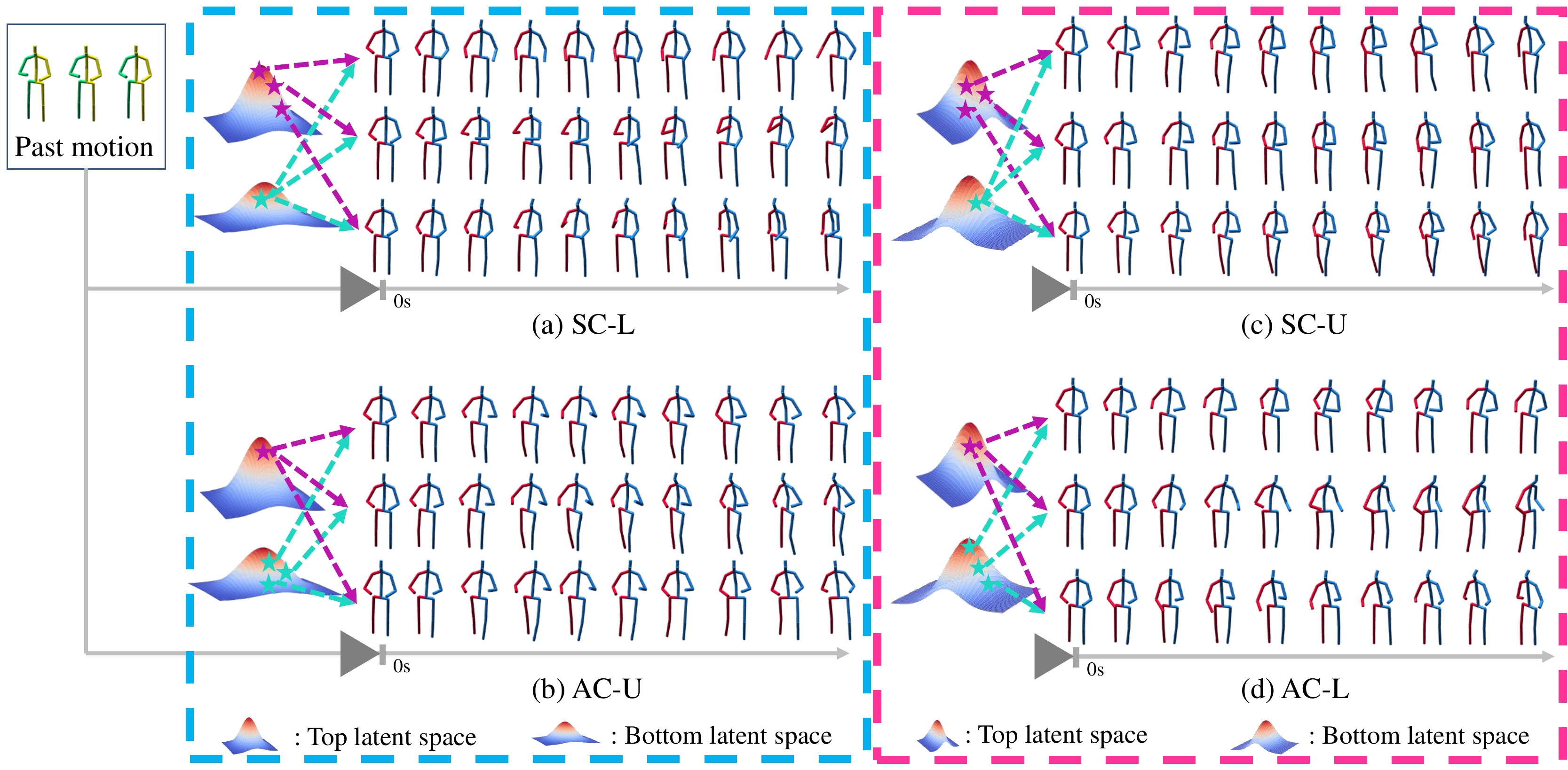}
\caption{Controllable human motion prediction. Given a past human motion, our model enables performing both strict control (SC) and adaptive control (AC) for the upper (U)/lower (L) partial body. In a single training, two types of control are provided for U/L body parts by fixing or varying the sampled latent variables from a pair of latent spaces, as shown in the blue (a,b) or magenta (c,d) box.}
\label{fig:teaser}
\end{figure*}

%
\IEEEpeerreviewmaketitle

\section{Introduction}
Human motion prediction is a fundamental problem in computer vision and has been broadly studied for decades. Modeling human motion can facilitate various kinds of applications in fields such as computer animation \cite{geijtenbeek2011interactive,van2010real} and antonymous driving \cite{ren2021safety,paden2016survey}. To solve the inherent ambiguity within human motion, recent work \cite{barsoum2018hp,yuan2020dlow,mao2021generating,yan2018mt,yuan2019diverse} focuses on modeling multimodal human motions, instead of the traditional deterministic regression for prediction \cite{martinez2017human,li2018convolutional,mao2019learning,mao2020history}.


Benefiting from the strong capacity of deep generative models, recent progress \cite{yuan2020dlow,mao2021generating} has even demonstrated predicting controllable human motions. The goal of controllable human motion prediction is to capture multiple predictions with intentionally fixed or varied partial-body motions. A simple way to control the predicted motions is to separately model each partial body using deep generative models (e.g., conditional variational auto-encoder (CVAE)). As such, the control can be realized by sampling latent codes from each latent space responsible for the corresponding partial body. However, this approach can induce unrealistic generations because it completely ignores the full-body kinematics and the intrinsic connection between each partial body. Alternatively, directly modeling full-body human motion ensures more reliable prediction, yet such a method does not straightforwardly lead to the capability to control the generation, as the latent space in this case can be heavily entangled with the mixture of a wide range of features. As a result, predicting both controllable and realistic human motions is challenging to achieve via a naive adoption of generative models. To better pursue controlled predictions, Yuan et al. \cite{yuan2020dlow} introduced deep mapping functions to seek desired motions with similar partial-body movements from a pre-trained CVAE that models full-body human motion. Toward designing an end-to-end controlling strategy, Mao et al. \cite{mao2021generating} proposed sequentially generating partial human bodies with a stochastic generation model. This method, however, only allows for varying body parts with a later generation order in the test stage. Considering the implicit relationship between parts of human body, state-of-the-art methods can be trapped into a dilemma of compromising learning valid human motion and realizing ideal control over each portion of body, making the models less capable of fulfilling both requirements in one training. 

In this paper, we introduce an end-to-end human motion modeling framework to facilitate control over desired body parts. We cast the problem of controllable human motion prediction as learning disentangled representations for generation, and propose to disentangle latent representations in the manner of imposing inductive bias for generation. Built on top of the common CVAE model (referred to as the top path) for learning full-body human motion, we couple in parallel an extra CVAE path (referred to as the bottom path) whose input is the motion for only a portion of the body (i.e., upper or lower body). A pair of latent variables (i.e., $\mathbf{z}_{t}$ for the top and $\mathbf{z}_{b}$ for the bottom) are involved for our network and are respectively endowed with specific generation roles for each partial human body.  In our formulation, $\mathbf{z}_b$ is encouraged to characterize the generation for one part of human body. As such, the disentangling mechanism then informs $\mathbf{z}_t$ with both the joint information for the other part of the human body, together with the implicit high-level information for inter-partial human body kinetics and relevance. Due to such a beneficial property, our strategy enables two types of control for each part of the body within a single training, which is hard for existing methods \cite{yuan2020dlow,mao2021generating} to achieve. Specifically, in generation, one can strictly control one portion of the body (e.g., lower body, Fig. \ref{fig:teaser}(a)) by generating with fixed $\mathbf{z}_b$; or predict adaptively controlled motions for the other portion by fixing  $\mathbf{z}_t$ (e.g., upper body, Fig. \ref{fig:teaser}(b)). The adaptive controlling fashion manifests itself mainly in adjusting some features 
(e.g., orientation) to cooperate with its given paired $\mathbf{z}_b$ to compose valid human motion, without losing the similarity in relative movements (e.g., all the hand movements are relatively similar with different degrees of yaw rotation in Fig. \ref{fig:teaser}(b)). The two controlling styles for the upper or lower body portion can be reversely realized with a second training by taking the other partial body as the bottom path input (e.g., Fig. \ref{fig:teaser}(a $\leftrightarrow$ c), Fig. \ref{fig:teaser}(b $\leftrightarrow$ d)). Also, flexibly customizing the bottom path input can enable other potential controlling manners. For example, we can control all the generated predictions to end up in a similar pose with various intermediate motions.

Additionally, stochastic human motion prediction usually calls for diversity in controlling, e.g., giving diverse upper-body motions with similar lower-body motions. To enrich the sample diversity, we extend and adapt the sampling strategy in \cite{yuan2020dlow} to our trained model to ensure a diverse mode coverage. In particular, we notice that in spite of a high average diversity, a mean-based diversity-promoting objective can lead to highly similar predictions \cite{mao2021generating,yuan2020dlow}. We thus leverage an alternative objective to promote duplication-aware sample diversity by penalizing the minimum pairwise distance between two samples. We also adopt the normalizing flow-based pose prior in \cite{mao2021generating} to mitigate sampling invalid poses. Benefiting from the disentangled latent spaces by our approach, training our sampler does not require any additional regularizer to explicitly constrain the generated partial body motions to be similar, which contributes to a task-specific optimization configuration for our sampler and hence yields a higher diversity.

Our main contributions consist of the following three aspects: (1) We propose a novel perspective to predict controllable human motion by formulating a disentanglement framework, which provides versatile control over generation via flexibly adjusting the network input; (2) we extend a sampling strategy to sample from our learned latent space to achieve non-repeated and controlled predictions with high sample diversity; (3) we present extensive experimental evaluations to demonstrate the state-of-the-art controllable human motion prediction performance of our model. 

\section{Related Work}
In this section, we first review previous work on stochastic human motion prediction, including controllable motion prediction methods. We then discuss some disentangled learning-based work in human motion modeling and image/video generation, since they share some similarities.  

\noindent	\textbf{Stochastic Human Motion Prediction.} 
The majority of previous attempts conduct deterministic sequence-to-sequence mapping for human motion prediction \cite{martinez2017human,mao2019learning}. These work typically enforces 
powerful network architectures to facilitate prediction, such as Graph Convolutional Networks (GCNs) \cite{mao2020history,dang2021msr,li2020dynamic} and Transformers \cite{cai2020learning,aksan2021spatio,martinez2021pose}. However, one prominent challenge in human motion is its inherent stochasticity, which makes the problem ill-posed even if past observations are given. As a result, later efforts switch to exploring stochastic approaches to produce multiple plausible predictions to model such inner ambiguity. Specifically, most of the stochastic prediction methods resort to deep generative models, such as condiational variational auto-encoders (CVAEs) \cite{aliakbarian2020stochastic,hassan2021stochastic,kania2021trajevae,aliakbarian2021contextually}, generative adversarial networks (GANs) \cite{barsoum2018hp,liu2021aggregated,wang2020learning} and normalizing flows (NFs) \cite{henter2020moglow}, to produce multiple predictions via repetitively sampling from the latent space.

Based on the impressive capability of stochastic models, recent work has been further developed towards predicting controllable human motion. Yuan et al. \cite{yuan2020dlow} presented a pioneering work on controlling partial-body movements. They leveraged a deep mapping network to search desired latent variables responsible for similar partial-body motions, from a pre-trained CVAE latent space for full-body human motion modeling. The mapping objective is also designed to include a diversity-promoting objective to guarantee the uncontrolled body parts  diversified. Yet, pursuing a rich sample diversity and a decent power of control simultaneously can limit the performance of the mapping network on both tasks. Mao et al. \cite{mao2021generating} directly employed a stochastic motion generation network and sequentially generated each body part to train the network end to end for control. This method provides absolute control because the earlier generated partial body is fixed and copied for all predictions, based on which the other parts are subsequently produced. However, as the order for sequential generation is predetermined, one cannot vary the body parts with an earlier generation order in the test stage. To the best our knowledge, these two work \cite{yuan2020dlow,mao2021generating} shares the closest motivation to ours in predicting controllable human motion, and constitutes the state of the art. Our method handles this problem from a different angle by learning to encode disentangled representations into latent spaces. Furthermore, our method, equipped with a diversity sampler, results in promising performance in both controlling strength and sample diversity.  

\begin{figure*}[t]
\centering
\includegraphics[width=1.0\linewidth]{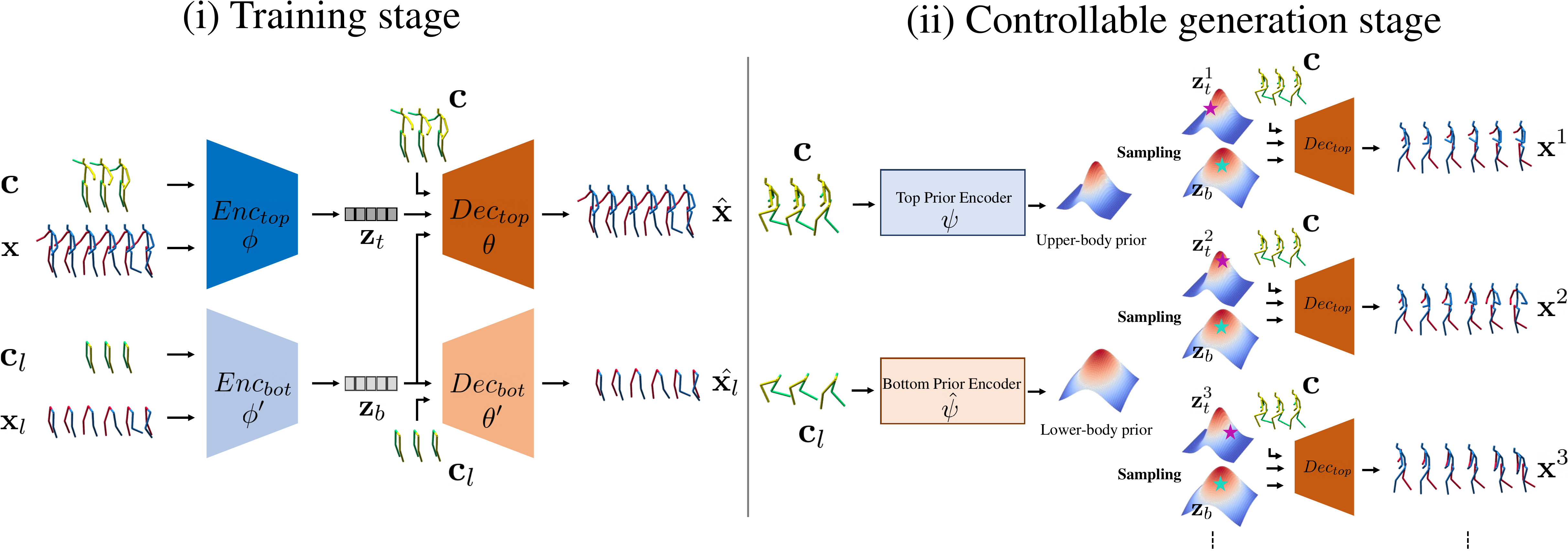}
\caption{Overview of our method. The past and future motion set for lower portion of the body $\begin{Bmatrix}\mathbf{c}_l, \mathbf{x}_l\end{Bmatrix}$ is assigned to the bottom path input as an example. After the end-to-end training in (i), the trained top path decoder enables producing controllable motion predictions in (ii). }
\label{fig:overview}
\end{figure*}

\noindent	\textbf{Disentanglement Learning.} 
Disentangling the latent space is another technique to obtain controlled generation. It is worth mentioning that disentanglement learning is more prevalent in image or video generation tasks \cite{guen2020disentangling,zhao2018recognize,nguyen2019hologan,vowels2021vdsm,skafte2019explicit}, but is relatively sparsely explored in human motion modeling. As mentioned in \cite{locatello2019challenging}, disentanglement fundamentally involves inductive bias. Zhao et al. \cite{zhao2017learning} proposed variational ladder auto-encoders to encode different levels of abstraction into a hierarchical structure for latent variables. Nguyen et al. \cite{nguyen2019hologan} imposed inductive bias toward explicit 3D rigid-body transformations to guide the generator to render images from multiple views. These work, however, generally requires iterative tuning on model architecture to achieve decent performance.
Later efforts \cite{charakorn2020explicit,jakab2018unsupervised,gu2021cgmvae} then attempt to elaborately design the inductive bias with involving the minimum change on the original generative model. Charakorn et al. \cite{charakorn2020explicit} introduced a SPLIT framework to globally and locally decouple the latent representations. Jakab et al. \cite{jakab2018unsupervised} stacked an additional key-point learning network to subtract from source and target image pairs the geometric information for disentanglement. Our method draws inspiration from these work in building a two-path network architecture but differs essentially in the usage and target as we focus on exploring versatile control over human motion prediction, which involves more complexity due to its sequential nature and intrinsic stochasticity.


In the scope of human motion modeling, studies \cite{yan2018mt,blattmann2021behavior,xu2020hierarchical,liu2021aggregated} have been conducted to control the synthesis of human dynamics under disentanglement frameworks. Yan et al. \cite{yan2018mt} directly performed latent space arithmetics to subtract and add the feature embedding to model the transformation for two motion modes. Blattmann et al. \cite{blattmann2021behavior} proposed to separate human behavior and posture with an auxiliary decoder to learn transition between motions. In essence, these work tackles controlling over global features and is designed for motion synthesis with feature transfer. Differently, our method concentrates on a finer level of control in terms of partial-body motions.

\section{Method}
We propose a disentanglement framework with a couple of CVAE paths for human motion prediction. We represent the observed human motion sequence as $\mathbf{c\\}=[x^{1},\dots,x^{h},\dots,x^{H}]$ of $H$ frames, in which $x^{h} \in \mathbb{R}^d$ denotes a human pose at the $h$-th frame with $d$ dimensions. Given past motion $\mathbf{c}$, we aim to predict the future motion $\mathbf{x}=[x^{H+1},x^{H+2},\dots,x^{H+T}]$ with $T$ frames. Our network architecture is illustrated in Fig. \ref{fig:overview}(i), which is trained end to end to learn disentangled representations for human motion. During training, our framework consists of two similar paths: 
top and bottom paths that respectively learn full-body and partial-body motion. Specifically, our trained top path decoder can be employed as a generator to predict controllable human motions, by taking as input the sampled latent codes from different latent spaces, as shown in Fig. \ref{fig:overview}(ii).

\subsection{Disentanglement for Controllable Motion Prediction}
\label{sec:method}
We start with a brief review of using CVAE for human motion modeling as our method involves CVAE for the network architecture. Then, we provide a detailed explanation of our disentanglement framework for controllable human motion prediction. Finally, to explore other applications in motion controlling with our proposed network architecture, we explain a different form of control that enables predicting multiple human motions ending with fixed poses.

\noindent	\textbf{Preliminary: CVAE for Human Motion Modeling.} CVAE is an extension of VAE that introduces an additional conditioning variable in the generative process. In the context of human motion prediction, CVAE learns to maximize the likelihood of the conditional data distribution $p(\mathbf{x}|\mathbf{c})$ for future motions $\mathbf{x} \in \mathcal{X}$, where $\cal{X}$ denotes the set of all possible future full-body motions. The condition is given by the observed sequence $\mathbf{c} \in \mathcal{C}$, where $\mathcal{C}$ represents a motion set containing all possible history observations. Such data distribution can be reparameterized by including a latent variable $\mathbf{z} \in \mathcal{Z}$, which can be expressed as $p(\mathbf{x}|\mathbf{c}) = \int p(\mathbf{x}|\mathbf{c},\mathbf{z})p(\mathbf{z})d\mathbf{z}$. In learning $p(\mathbf{x}|\mathbf{c},\mathbf{z})$, CVAE utilizes an  $\phi$-parameterized encoder network to embed $\mathbf{x}$ together with $\mathbf{c}$ into the latent variable $\mathbf{z}$, and then reconstruct it using a decoder network with parameters $\theta$. Conventionally, due to the mathematical tractability and sample simplicity, the posterior $q_{\phi}(\mathbf{z}|\mathbf{x},\mathbf{c})$ is regularized to match a Gaussian prior $\mathcal{N}(0,I)$ in the network optimization. The trained CVAE decoder can then be exploited for predicting multimodal human motion by feeding it with $\mathbf{c}$ and sampled $\mathbf{z}$, following:
\begin{eqnarray}
\label{eq:eq1}
\mathbf{z} \sim p(\mathbf{z}),\\
\label{eq:eq2}
\mathbf{x} \sim p_{\theta}(\mathbf{x}|\mathbf{c},\mathbf{z}).
\end{eqnarray}
However, sampling randomly from the latent space does not ideally lead to controlled predictions because a sampled latent code commonly entails entangled features for human motion.


\noindent	\textbf{Disentangling Human Motion.} Let us now detail our disentanglement framework for controllable human motion prediction. Formally, we split the past and future human motion into two parts: $\mathbf{c} = [\mathbf{c}^{1}_{p}, \mathbf{c}^{2}_{p}], \mathbf{x} = [\mathbf{x}^{1}_{p}, \mathbf{x}^{2}_{p}]$, where $\mathbf{c}^{i}_{p} \in \mathbb{R}^{H \times d_i}$ and $\mathbf{x}^{i}_{p} \in \mathbb{R}^{T \times d_i}$  $(i=1,2)$ respectively define the history and future motion for one body portion (i.e, upper or lower body) that consists of $d_i$-dimensional human poses. Our goal is to encode different representations into separate latent variables and let each of them govern different features for generation. We thus assume the partial-body motion $\mathbf{x}^{1}_p$ is generated via latent variable $\mathbf{z}_b$, and the full-body human motion $\mathbf{x}$ (including both portions of the body) requires the latent variable pair ($\mathbf{z}_t$, $\mathbf{z}_b$) for generation. To facilitate a two-fold controlling capability with respect to both portions of the body in generation, we model $\mathbf{z}_
t$ and $\mathbf{z}_b$ to be independent, and assume the following new generative process: 
\begin{eqnarray}
\label{eq:eq3}
\mathbf{z}_b \sim p(\mathbf{z}_b),\\
\label{eq:eq4}
\mathbf{z}_t \sim p(\mathbf{z}_t),\\
\label{eq:eq5}
\mathbf{x} \sim p(\mathbf{x}|\mathbf{z}_t,\mathbf{z}_b,\mathbf{c}),\\
\label{eq:eq6}
\mathbf{x}^{1}_p \sim p(\mathbf{x}^{1}_p|\mathbf{z}_b,\mathbf{c}^{1}_p),
\end{eqnarray}
in which the conditioning variables $\mathbf{c}$ and $\mathbf{c}^{1}_p$ are included to meet the task of motion prediction. We respectively parameterize the generator for $\mathbf{x}$ and $\mathbf{x}^{1}_p$ with ${\theta}$ and 
${\theta}^{\prime}$, and adopt the CVAE learning scheme by introducing two encoder networks to optimize the mappings ${\theta}$ and 
${\theta}^{\prime}$.  Straightforwardly, our model can be understood as two CVAE paths: (1) The first CVAE (the top path) that models full-body human motion $\mathbf{x}$ conditioned on $\mathbf{c}$; (2) the extra path (the bottom path) that learns only the corresponding partial-body motion $\mathbf{x}^{1}_p$  conditioned on $\mathbf{c}^{1}_p$. To fulfill the above generative process, we concatenate the bottom path latent variable $\mathbf{z}_b$ with the top path one $\mathbf{z}_t$, and feed them into the top path decoder together with the condition $\mathbf{c}$ to generate the full-body human motion.

The key advantage of our modeling lies in that the existence of the bottom path plays the role of a strong inductive bias, forcing $\mathbf{z}_t$ to avoid the repeated information in $\mathbf{z}_b$ such that the representations for $\mathbf{z}_t$ and $\mathbf{z}_b$ do not overlap. With the bottom path embedding all the partial-body information for $\mathbf{x}^{1}_p$ into $\mathbf{z}_b$, 
the disentangling mechanism thus motivates $\mathbf{z}_t$ to focus only on the remaining information except for the representations for $\mathbf{x}^{1}_p$. Since the original target of the top path is to learn the full-body human motion, the implicit high-level representations responsible for partial-body relevance and motion validity (e.g., kinematics) must also be modeled in the top latent space. As stated in the generative process, $\mathbf{z}_t$ and $\mathbf{z}_b$ are expected to jointly entail all the information to form a valid and realistic full-body human motion. As such, the representations within $\mathbf{z}_t$ need to convey more than the joint coordinate information for the other body portion $\mathbf{x}^{2}_p$. More precisely, $\mathbf{z}_t$ further captures the necessary implicit relationship between each body part (i.e., $\mathbf{x}^{1}_p$ and $\mathbf{x}^{2}_p$) to compose valid human motion because these representations are not included in $\mathbf{z}_b$ in our framework.

This property conveniently enables our trained top path decoder to show two types of controlling style in the test stage. In the case of the bottom path input $\mathbf{x}^{1}_p$ being the motion for lower human body, a single training for our model can yield:  \textit{Type(1)}: strictly fixed lower partial-body with different upper-body motions by sampling fixed $\mathbf{z}_b$ and varied $\mathbf{z}_t$; \textit{Type(2)}: different lower-body motions and adaptively controlled upper-body motions in which some features are automatically adjusted (e.g., orientation) to cooperate with different lower-bodies but still maintains similar relative upper-body motions, by sampling fixed $\mathbf{z}_t$ and varied $\mathbf{z}_b$. Such a difference stems from the additional information for partial-body relevance in $\mathbf{z}_t$, which empowers our strategy to a show clear superiority from naively learning each partial-body motion individually for generation.

\noindent	\textbf{Optimization.} Following the generative process in Eq. \ref{eq:eq3} $\sim$ 
Eq. \ref{eq:eq6}, our network can then be optimized via \textit{maximizing} the following Evidence Lower Bound (derivation is shown in the APPENDIX B):
\begin{equation}
\begin{aligned}
\label{eq:eq7}
L = &\underbrace{\mathbb{E}_{q_{\phi}(\mathbf{z}_t|\mathbf{x},\mathbf{c}), q_{{\phi}^{\prime}}(\mathbf{z}_b|\mathbf{x}^{1}_p,\mathbf{c}^{1}_p)}[\log p_{\theta}(\mathbf{x}|\mathbf{z}_t, \mathbf{z}_b, \mathbf{c})]}_{L^{top}_{rec}} \\+ & \underbrace{\mathbb{E}_{q_{{\phi}^{\prime}}(\mathbf{z}_b|\mathbf{x}^{1}_p,\mathbf{c}^{1}_p)}[\log p_{{\theta}^{\prime}}(\mathbf{x}^{1}_p|\mathbf{z}_b, \mathbf{c}^{1}_p)]}_{L^{bot}_{rec}} \\- &  \underbrace{\mathrm{KL}(q_{\phi}(\mathbf{z}_t|\mathbf{x},\mathbf{c})||p(\mathbf{z}_t))}_{L^{top}_{\mathrm{KL}}} - \underbrace{\mathrm{KL}(q_{{\phi}^{\prime}}(\mathbf{z}_b|\mathbf{x}^{1}_p,\mathbf{c}^{1}_p)||p(\mathbf{z}_b))}_{L^{bot}_{\mathrm{KL}}} ,
\end{aligned}
\end{equation}
in which $\phi$ and ${\phi}^{\prime}$ denote the encoder parameters in the top and bottom path, respectively. In Eq. \ref{eq:eq7}, two reconstruction terms and two Kullback-Leibler ($\mathrm{KL}$) terms are involved for the corresponding top and bottom path. Two reconstruction terms are implemented with Mean Square Error (MSE) between the input and the decoded (indicated with $\hat{\cdot}$) motions to measure the reconstructing error: 
\begin{eqnarray}
\label{eq:eq8}
L^{top}_{rec}  = \norm{\mathbf{x}-\hat{\mathbf{x}}}^2 ,\\
\label{eq:eq9}
L^{bot}_{rec} = \norm{\mathbf{x}^{1}_p-\hat{\mathbf{x}^{1}_p}}^2.
\end{eqnarray}

Existing work \cite{zhang2021we,yuan2020dlow} commonly applies a standard Gaussian prior $\mathcal{N}(0,I)$ to regularize the latent space. As pointed out in \cite{aliakbarian2021contextually}, using such a condition-free prior for generation could lead to the loss of context information. To show awareness of the context in the generation stage, we make the prior learnable by mapping the condition to Gaussian parameters (i.e., mean and standard deviation). The KL terms can thus be evaluated as
\begin{eqnarray}
\label{eq:eq10}
L^{top}_{\mathrm{KL}} =  \mathrm{KL}(q_{\phi}(\mathbf{z}_t|\mathbf{x},\mathbf{c})||p_{\psi}(\mathbf{z}_t|\mathbf{c})),\\
\label{eq:eq11}
L^{bot}_{\mathrm{KL}} =  \mathrm{KL}(q_{{\phi}^{\prime}}(\mathbf{z}_b|\mathbf{x}^{1}_p,\mathbf{c}^{1}_p)||p_{{\psi}^{\prime}}(\mathbf{z}_b|\mathbf{c}^{1}_p)),
\end{eqnarray}
in which $\psi$ and ${\psi}^{\prime}$ are the parameters for the top and bottom prior encoding networks, respectively. We then rewrite the generative process in Eq. \ref{eq:eq3} $\sim$ Eq. \ref{eq:eq5} to reflect our learnable priors for the generation of $\mathbf{x}$: 
\begin{eqnarray}
\label{eq:eq12}
\mathbf{z}_t \sim p_{\psi}(\mathbf{z}_t|\mathbf{c}),\\
\label{eq:eq13}
\mathbf{z}_b \sim p_{{\psi}^{\prime}}(\mathbf{z}_b|\mathbf{c}^{1}_p),\\
\label{eq:eq14}
\mathbf{x} \sim p_{\theta}(\mathbf{x}|\mathbf{z}_t,\mathbf{z}_b,\mathbf{c}).
\end{eqnarray}
Eventually, the overall optimization objective to \textit{minimize} can be given by:
\begin{equation}
\label{eq:eq15}
\begin{aligned}
L &= \norm{\mathbf{x}-\hat{\mathbf{x}}}^2  + \norm{\mathbf{x}^{1}_p-\hat{\mathbf{x}^{1}_p}}^2 + \lambda^{t}_{\mathrm{KL}}\mathrm{KL}(q_{\phi}(\mathbf{z}_t|\mathbf{x},\mathbf{c})||p_{\psi}(\mathbf{z}_t|\mathbf{c}))\\
&+ \lambda^{b}_{\mathrm{KL}} \mathrm{KL}(q_{{\phi}^{\prime}}(\mathbf{z}_b|\mathbf{x}^{1}_p,\mathbf{c}^{1}_p)||p_{{\psi}^{\prime}}(\mathbf{z}_b|\mathbf{c}^{1}_p)),
\end{aligned}
\end{equation}
in which the $\lambda^{t}_{\mathrm{KL}}$ and  $\lambda^{b}_{\mathrm{KL}}$ are weighted to the $\mathrm{KL}$ terms to balance the quality of reconstruction and latent space, following \cite{higgins2017beta}. Our network (i.e., both paths) can be end-to-end optimized. We can then yield controlled generation with the trained top path decoder. 

\begin{figure*}[t]
\centering
\includegraphics[width=0.8\linewidth]{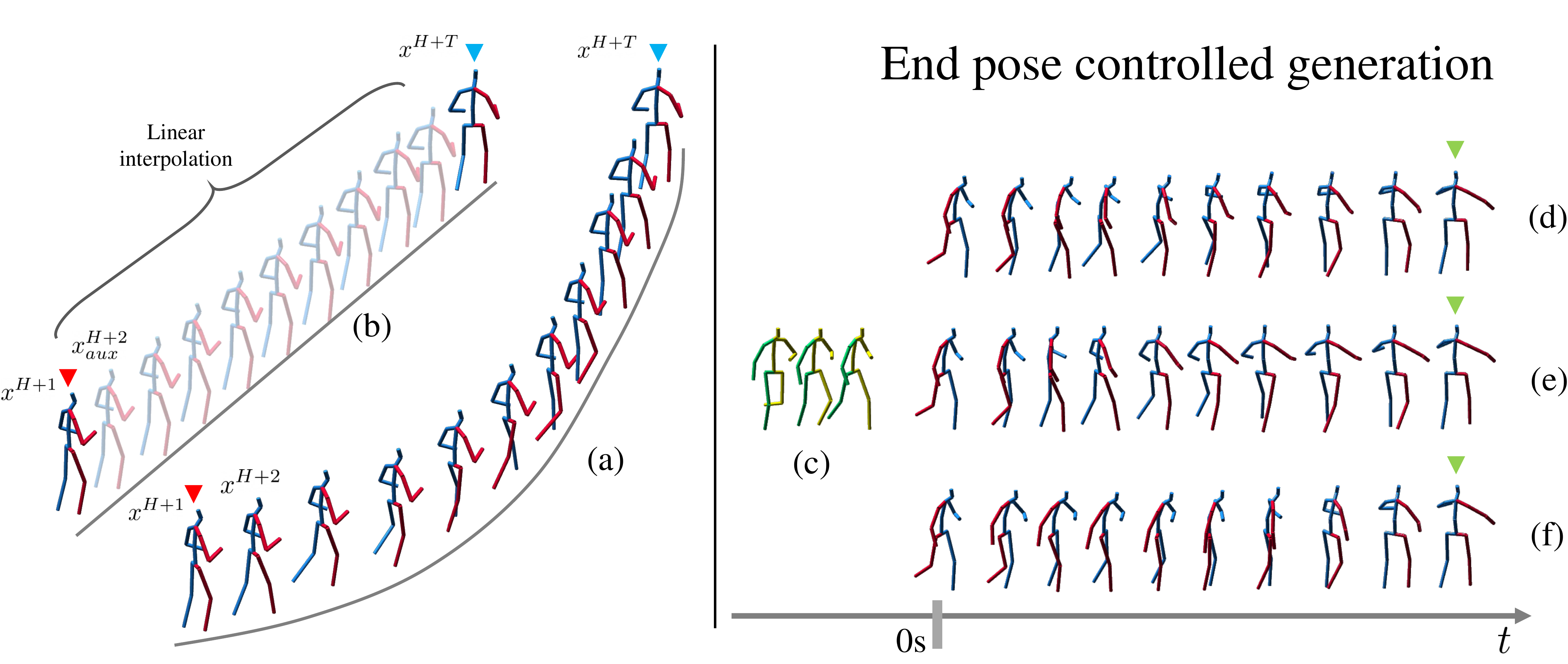}
\caption{An example of obtaining the bottom path input (left) for end pose controlled human motion perdition (right). The transparent poses in (b) are linearly interpolated based on the start (red arrows) an end (blue arrows) poses from the original sequence (a). In generation, given past motion (c), our model predicts multiple sequences (d) $\sim$ (f) with varied intermediates motions, but the same final poses (green arrows). }
\label{fig:end_pose_fig}
\end{figure*}

\noindent	\textbf{End Pose Controlled Human Motion Prediction.}
The core insight of imposing inductive bias via our formulation can be further extended to disentangle different forms of representations that contribute to generating the human motion $\mathbf{x}$ by flexibly adjusting the bottom path input. In human motion modeling, a possible control type would be generating multiple motion predictions that end up in a similar pose, but with different intermediate motions to reach such an end pose. To realize this, our target is now switched to forcing $\mathbf{z}_t$ to remove the representations that control the end pose and instead let $\mathbf{z}_b$ govern it. We therefore create an auxiliary sequence $\mathbf{x}_{aux}$ that ends up in the same final pose with $\mathbf{x}$ to construct the required inductive bias. With the same input $\begin{Bmatrix}\mathbf{c},\mathbf{x}\end{Bmatrix}$ for the top path, the bottom path input is now designed as $\begin{Bmatrix}\mathbf{c},\mathbf{x}_{aux}\end{Bmatrix}$. As such, $\mathbf{z}_t$ is encouraged to avoid representing the repeated information with respect to the end pose, and concentrate only on the intermediate motions. As illustrated in Fig. \ref{fig:end_pose_fig} (left), to inject as many variations as possible into $\mathbf{z}_t$, the input of the bottom path $\mathbf{x}_{aux}$ is designed to be a linearly interpolated sequence from the start ${x}^{H+1}$ to end frame ${x}^{H+T}$ to reach the same length with $\mathbf{x}$. $\mathbf{x}_{aux}$ can then be expressed as $\mathbf{x}_{aux} = [{x}^{H+1}, {x}^{H+2}_{aux}, {x}^{H+3}_{aux},..., {x}^{H+T-1}_{aux}, {x}^{H+T}]$ including $T-2$ interpolated poses $[{x}^{H+2}_{aux}, \dots,{x}^{H+T-1}_{aux}$]. Essentially, the linear interpolation leads $\mathbf{x}_{aux}$ to reflect a smooth pose sequence whose motion complexity is intentionally destroyed  compared with $\mathbf{x}$. Thus, $\mathbf{x}_{aux}$ now contains the minimum amount of movements between ${x}^{H+1}$ and ${x}^{H+T}$, which leaves more intermediate motion variations in $\mathbf{z}_t$. We then redefine Eq. \ref{eq:eq15} to express the new training objective as
\begin{equation}
\label{eq:eq16}
\begin{aligned}
L &= \norm{\mathbf{x}-\hat{\mathbf{x}}}^2  + \norm{\mathbf{x}_{aux} - \hat{\mathbf{x}_{aux}}}^2 \\ &+ \lambda^t_{\mathrm{KL}}\mathrm{KL}(q_{\phi}(\mathbf{z}_t|\mathbf{x},\mathbf{c})||p_{\psi}(\mathbf{z}_t|\mathbf{c})) \\
&+ \lambda^b_{\mathrm{KL}} \mathrm{KL}(q_{{\phi}^{\prime}}(\mathbf{z}_b|\mathbf{x}_{aux},\mathbf{c})||p_{{\psi}^{\prime}}(\mathbf{z}_b|\mathbf{c})).
\end{aligned}
\end{equation}

Similarly, in the generation stage with the trained top path decoder, we can pick a certain end pose by fixing $\mathbf{z}_b$, and then generate end pose controlled predictions with multiple intermediate motions by varying $\mathbf{z}_t$ (Fig. \ref{fig:end_pose_fig}, right). 

\noindent	\subsection{Diversity Sampling from Disentangled Latent Spaces}
Although our model enables predicting controllable human motions, the simple random sampling does not ensure sample diversity in generation. To alleviate this problem, motivated by \cite{zhang2021we,cui2021lookout,yuan2021agentformer}, we extend the sampler Dlow \cite{yuan2020dlow} to cover diverse modes of predictions from our learned latent spaces.
In particular, Dlow learns to map a sampled noise $\epsilon \sim \mathcal{N}(0,I)$ to a set of latent variables $\begin{Bmatrix} \mathbf{z}^{1},\dots,\mathbf{z}^{k},\dots,\mathbf{z}^{K} \end{Bmatrix}$ whose decoded prediction set $\begin{Bmatrix}\mathbf{x}^{1},\dots,\mathbf{x}^{k},\dots,\mathbf{x}^{K}\end{Bmatrix}$ is richly diversified. The mapping network parameterized by $\eta$ takes the condition $\mathbf{c}$ as input and linearly transforms $\epsilon$ such that $\mathbf{z}^{k}$ remains Gaussian distributed $r_{\eta}(\mathbf{z}^{k}|\mathbf{C})$ to circumvent the domain shift:
\begin{equation}
\label{eq:eq17}
\mathbf{z}^{k} = \mathbf{A}^{k}\epsilon + \mathbf{b}^{k}.
\end{equation}
$\mathbf{A}^{k} \in \mathbb{R}^{d_z}$ and $\mathbf{b}^{k} \in \mathbb{R}^{d_z}$ are two vectors with the same number of dimension ${d_z}$ as $\mathbf{z}^{k}$. The resulting loss for the sampler involves a $\mathrm{KL}$ loss to regularize the mapped latent variables to remain Gaussian and a diversity loss formulated with the mean-based pairwise distance of predicted motions. Since we enforce learnable prior $p(\mathbf{z}|\mathbf{c})$, the $\mathrm{KL}$ loss for our sampler can be defined as
\begin{equation}
\label{eq:eq18}
L^{samp}_{\mathrm{KL}}  = \sum_{k=1}^{K} \mathrm{KL}( r_{\eta}(\mathbf{z}^{k}|\mathbf{c}) || p(\mathbf{z}|\mathbf{c})).
\end{equation}

\noindent	\textbf{Diversity Objective.} Despite the achieved high average diversity among all the predictions, we have noticed that the mean-based diversity objective can lead to trivial solutions that several generated samples can show huge resemblance. To relieve this issue, inspired by \cite{ma2021likelihood}, we instead penalize the minimum pairwise $\mathcal{L}$2 distance to promote diversified predictions with low repetition:
\begin{equation}
\label{eq:eq19}
L_{div} = \underset{i \neq j \in  \lbrace1,\dots,K\rbrace}{\min}\norm{\mathbf{x}^{i}-\mathbf{x}^{j}}^2.
\end{equation}
Differently from the 2D trajectory forecasting task in \cite{ma2021likelihood} that only diversifies the predicted endpoint, our diversity objective (Eq. \ref{eq:eq19}) guides the sampler to show ample variations across all the motion sequence pairs. 

\noindent	\textbf{Validity Objective.} Additionally, as argued in \cite{mao2021generating}, Dlow does not pay attention to the potentially invalid poses in the latent space, leading to the possibility of sampling diverse motions in which some poses are unreliable. We thus adopt a normalizing flow-based pose prior in \cite{mao2021generating} as a validity promoter to combat the tendency of sampling unrealistic poses. Normalizing flow \cite{henter2020moglow,rezende2015variational} aims to learn a mapping function $f$, which is parameterized with deep neural networks, to minimize the negative log-likelihood of the samples. The trained network $f^{*}$ can then be utilized as a validity objective $L_{vli}$ to motivate the sampler to avoid invalid poses. A detailed explanation is provided in the APPENDIX B.

We then compose Eq. \ref{eq:eq18}, Eq. \ref{eq:eq19}, and the validity objective $L_{vli}$ to construct the final loss to train our sampler. It is noteworthy that our model naturally provides control for each body portion due to the disentangled representations. Therefore, we can perform diversity sampling from one latent space but keep the sampled latent variable from the other latent space fixed to make the predictions both controllable and diverse among different body parts. This results in different losses to guide the sampler. In the context of generating predictions with diverse upper-body motions $\mathbf{x}_u$ but controlled lower-body motions, the sampler loss can be expressed as
\begin{equation}
\label{eq:eq20}
\begin{aligned}
L_{samp} = &\lambda^{s}_{\mathrm{KL}}L^{samp}_{\mathrm{KL}}  + \lambda_{div} L_{div} + \lambda_{vli}L_{vli} \\
= &\lambda^{s}_{\mathrm{KL}}\sum_{k=1}^{K} \mathrm{KL}( r_{\eta}(\mathbf{z}^{k}_t|\mathbf{c}) || p_{\psi}(\mathbf{z}_t|\mathbf{c})) \\ +& \lambda_{div}\underset{i \neq j \in \lbrace1,\dots,K\rbrace}{\min}\norm{\mathbf{x}^{i}_u-\mathbf{x}^{j}_u}^2 + \lambda_{vli}L_{vli},
\end{aligned}
\end{equation}
where $\lambda^{s}_{\mathrm{KL}}$, $\lambda_{div}$ and $\lambda_{vli}$ are the weights for the corresponding objective.  


\section{Experiment}

\noindent	\textbf{Datasets.} Following \cite{yuan2020dlow,mao2021generating}, we assess our method on a benchmark dataset: Human3.6M \cite{ionescu2013human3}. Human3.6M is a large human motion dataset with 15 daily scenarios performed in seven subjects recorded at 50 Hz. We use S1, S5, S6, S7, and S8 for training, and test on S9 and S11. We adopt a 17-joint human skeleton, and predict the future 100 frames (two seconds) based on the given 25 history frames (0.5 seconds). The global translation of person is removed. 

\noindent	\textbf{Network and Implementation Details.}
All the encoders and decoders in our method are construed under Gated Recurrent Unit (GRU) \cite{cho2014learning} considering the sequential nature of human motion data. We set the dimension $d_z$ of both latent variables (i.e., $\mathbf{z}_t$ and $\mathbf{z}_b$) to 128, 
and train our model for 500 epochs on RTX 3090 using the ADAM optimizer \cite{kingma2014adam}. We empirically set $(\lambda^{t}_{\mathrm{KL}}, \lambda^{b}_{\mathrm{KL}})$ to $(0.1, 0.1)$ for our model and $(\lambda^{s}_{\mathrm{KL}}, \lambda_{div}, \lambda_{vli})$ to $(1.0, 0.7, 0.7)$ for the diversity sampler. We clip the value of $L_{div}$ to $[0, 160]$ to prevent it from diverging. Our sampler is trained for 100 epochs with the learning rate of 1e-4 using ADAM.

\begin{figure*}[!t]
\centering
\includegraphics[width=0.84\linewidth]{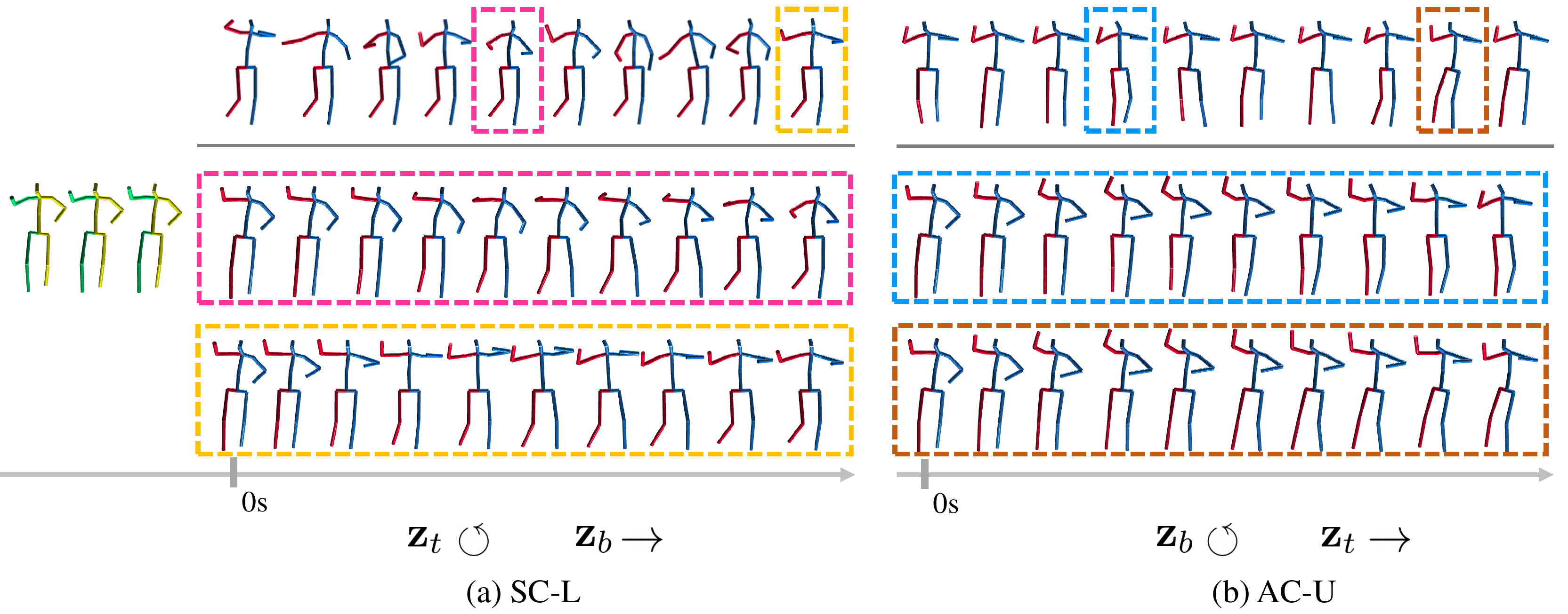}
\caption{Results of controllable human motion prediction via fixing (``$\rightarrow$") or varying (``$\circlearrowleft$") $\mathbf{z}_t$ and $\mathbf{z}_b$, with the bottom path input (BPI) being the lower body portion $\mathbf{x}_l$ in training. The top row shows the end poses of 10 generated sequences. Below we present different frames for two boxed samples.  }
\label{fig:res1}
\end{figure*}

\begin{figure*}[!t]
\centering
\includegraphics[width=0.84\linewidth]{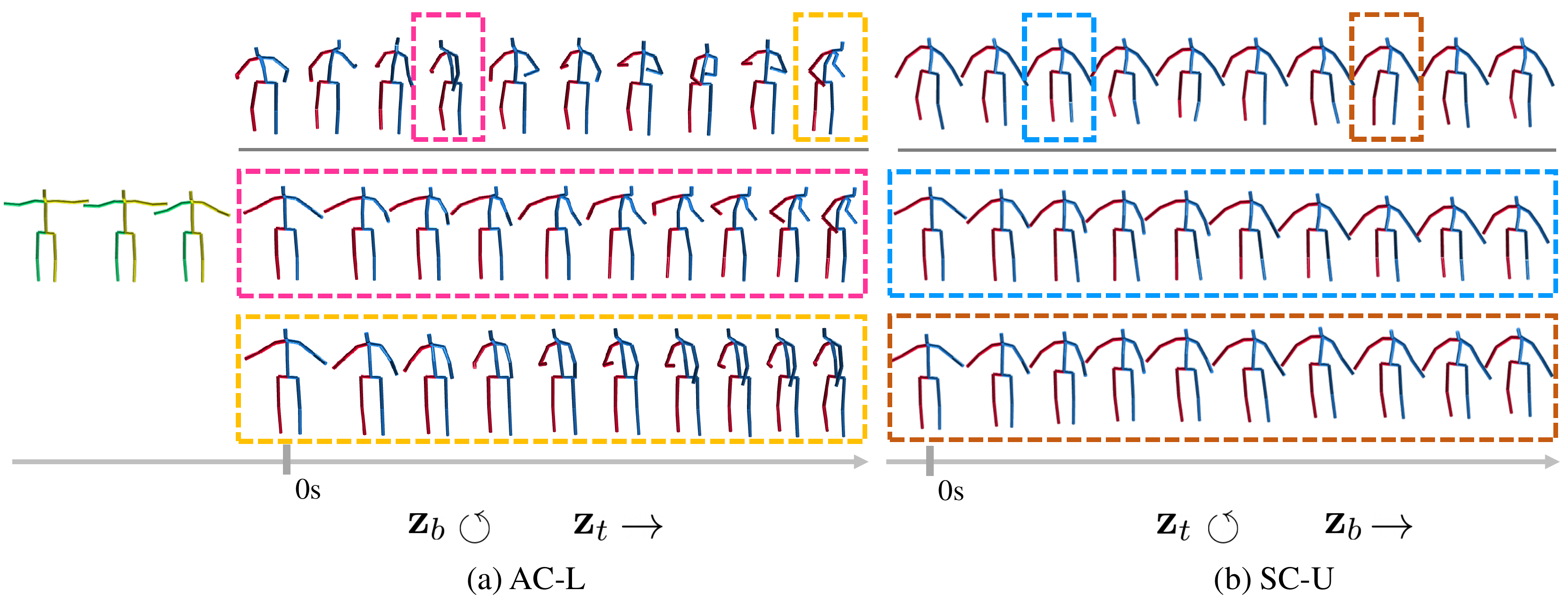}
\caption{Results of controllable human motion prediction via fixing (``$\rightarrow$") or varying (``$\circlearrowleft$") $\mathbf{z}_t$ and $\mathbf{z}_b$, with the BPI being the upper body portion $\mathbf{x}_u$ in training.}
\label{fig:res2}
\end{figure*}

\begin{table*}
\centering
\caption
{Evaluation of the controlling performance via the APD metric. Fixing and varying are indicated as ``$\rightarrow$" and ``$\circlearrowleft$", respectively. The 3rd $\sim$ 6th columns show the influence of different BPIs on the controlling ability.   }
\label{tab:tab1}
\begin{tabular}{cccccc}
\toprule
                     & CVAE                 & \specialcell{Ours (BPI: $\mathbf{x}_l$)\\$\mathbf{z}_t \circlearrowleft$ , $\mathbf{z}_b \rightarrow$ \\ SC-L }                 &\specialcell{Ours (BPI: $\mathbf{x}_l$)\\$\mathbf{z}_b \circlearrowleft$ , $\mathbf{z}_t \rightarrow$ \\ AC-U}                 & \specialcell{Ours (BPI: $\mathbf{x}_u$)\\$\mathbf{z}_t \circlearrowleft$ , $\mathbf{z}_b \rightarrow$\\ SC-U}               & \specialcell{Ours (BPI: $\mathbf{x}_u$)\\$\mathbf{z}_b \circlearrowleft$ , $\mathbf{z}_t \rightarrow$ \\ AC-L}                 \\ \hline
APD (Upper)          & 6.051                & 5.782                & 0.972                & 0.787                & 5.670                \\
APD (Lower)          & 2.958                & 0.392                & 3.047                & 2.515                & 0.813                \\ \bottomrule
\multicolumn{1}{l}{} & \multicolumn{1}{l}{} & \multicolumn{1}{l}{} & \multicolumn{1}{l}{} & \multicolumn{1}{l}{} & \multicolumn{1}{l}{}
\end{tabular}
\end{table*}

\begin{table*}
\centering
\caption
{Evaluation of the controlling performance and sample diversity. The results of BoM \cite{bhattacharyya2018accurate} and Dlow \cite{yuan2020dlow} (w/RS) are reported directly from \cite{mao2021generating}. ``Dlow RS" denotes adding a rejection sampling (RS) for Dlow, as suggested in \cite{mao2021generating}.}
\label{tab:tab2}
\begin{tabular}{cccccc}
\toprule
            & BoM \cite{bhattacharyya2018accurate}        & 
            \specialcell{ Dlow \cite{yuan2020dlow}\\ RS}  & Dlow \cite{yuan2020dlow}   & Gsps \cite{mao2021generating}        & Ours            \\ \hline
APD (Upper) $\uparrow$ & 4.408      & 7.280 & 12.741 & 13.150     & \textbf{15.426} \\
APD (Lower) $\downarrow$& \textbf{0} & 0.780 & 1.071  & \textbf{0} & 0.736           \\
MPD     $\uparrow$    & -          & -     & 4.446  & 1.654      & \textbf{12.810} \\ \bottomrule
\end{tabular}
\end{table*}

\begin{figure*}
\centering
\includegraphics[width=0.8\linewidth]{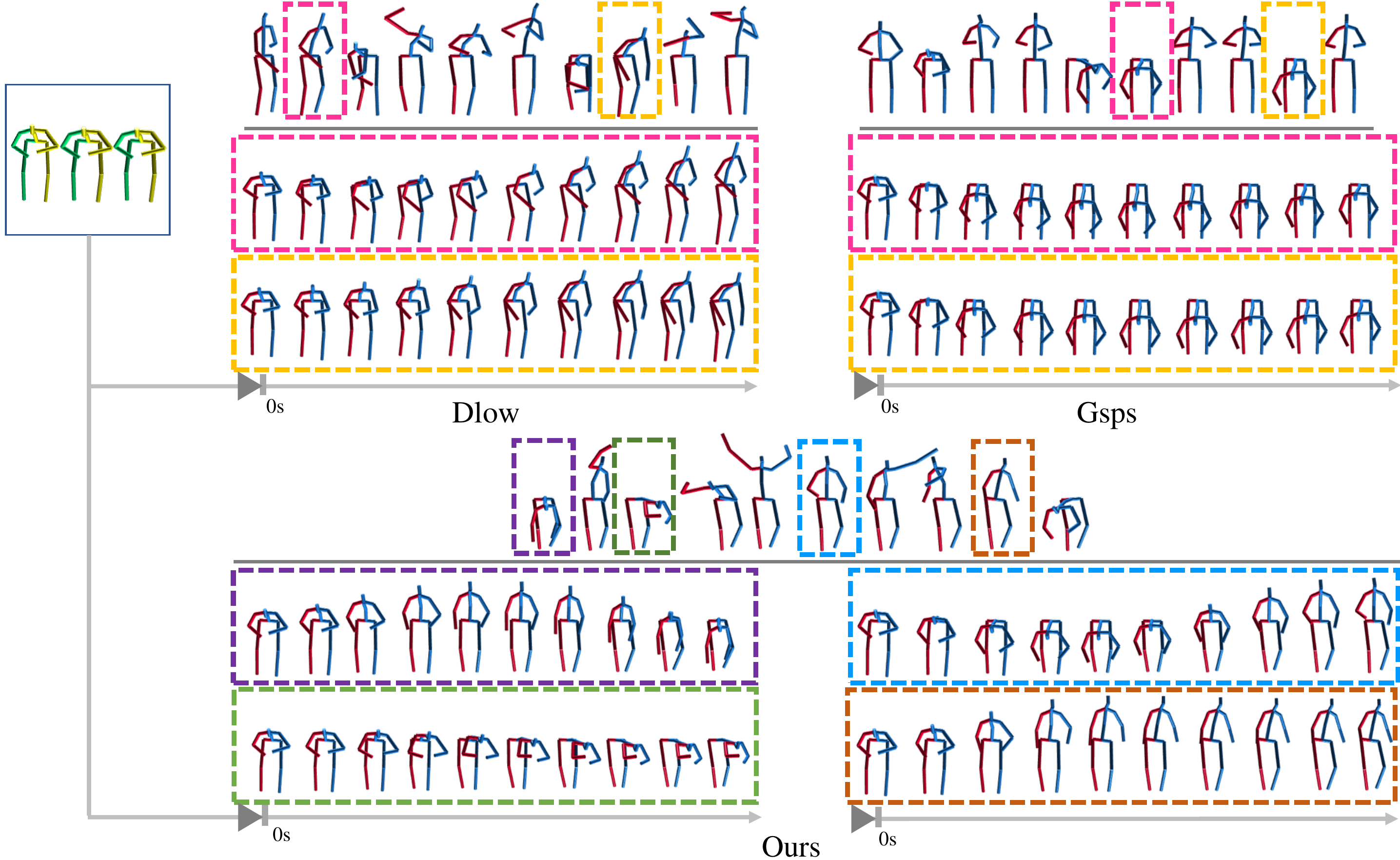}
\caption{Results of diversified controllable human motion prediction. All the methods aim to produce diverse upper-body motions with strictly fixed lower-body motions. For each method, the first row exhibits the end pose of 10 predictions. Below we present different frames for the boxed samples that end similarly for further comparison. Best view in color. } 
\label{fig:res3}
\end{figure*}

\subsection{Evaluation of Partial Body Controlled Prediction}
\noindent \textbf{Controllable Prediction.} We here investigate the controlling ability of our model in terms of partial body by generating motion predictions using the trained top path decoder, with \textit{randomly sampled} the latent variable pair ($\mathbf{z}_t,\mathbf{z}_b$). We first qualitatively report the controlling performance with respect to different bottom path input (BPI) selections. Fig. \ref{fig:res1} and Fig. \ref{fig:res2} visualize the generated predictions via training our model with lower-body motions $\mathbf{x}_l$ and upper-body motions $\mathbf{x}_u$ as the BPIs, respectively. It can be seen in Fig. \ref{fig:res1}(a) that in our model, the controlling power of fixing $\mathbf{z}_b$ leads to strictly fixed lower-body motions among all the generated samples, whereas fixing $\mathbf{z}_t$ contributes to a more adaptive manner of control relying on the lower-body motions. For example, with the lower bodies walking in different directions in Fig. \ref{fig:res1}(b), the upper bodies also turn correspondingly to make the motions realistic and natural. In spite of the orientation change, the relative movements for the upper bodies in Fig. \ref{fig:res1}(b), such as hand motions, remain the same. The key insight behind this is that $\mathbf{z}_t$ entails information for both the upper-body joint location and the unseen high-level representations between each body part for motion realism. Similarly, taking upper-body motions for the BPI leads to opposite roles for $\mathbf{z}_b$ and $\mathbf{z}_t$, as displayed in Fig. \ref{fig:res2}.

We further quantitatively show the controlling ability of our model via the Average Pairwise Distance (APD) metric \cite{yuan2020dlow}: $\frac{1}{K(K-1)}\sum_{i=1}^K \sum_{j=1, j\neq i}^K||\mathbf{x}^{i}-\mathbf{x}^{j}||_2$. APD measures the average $\mathcal{L}$2 distance between all the prediction pairs to evaluate diversity. A lower APD indicates less variations among generated predictions, demonstrating a higher controlling power. The number of the generated samples $K$ is set to 50. As shown in Tab. \ref{tab:tab1}, the selected body portion for the BPI results in a strict controlling power over it in generation (3rd and 5th columns), while making the control over the other half of the body less strict (4th and last columns). Additionally, a simple CVAE modeling for full-body motion hardly enjoys control. We can thus verify via the qualitative  (Fig. \ref{fig:res1}, \ref{fig:res2}) and the quantitative results (Tab. \ref{tab:tab1}) that $\mathbf{z}_t$ and $\mathbf{z}_b$ possess different types of strength in controlling generation, depending on the body portion chosen for BPI.

\begin{table}
\centering
\caption
{Quantitative evaluation of the influence of $L_{vli}$ on pose quality (NLL) and motion diversity (APD). }
\label{tab:tab3}
\begin{tabular}{cccccccc}
\toprule
                     &                      & Gsps \cite{mao2021generating}                 &                      & Ours (w/o $L_{vli}$)       &                      & Ours (w/ $L_{vli}$)        &                      \\ \cline{1-8}
NLL $\downarrow$                 &                      & 75.150               &                      & 76.261               &                      & \textbf{70.513}      &                      \\
APD $\uparrow$                 &                      & 13.150                    &                      & \textbf{15.628}      &                      & 15.426               &                      \\ \bottomrule
\multicolumn{1}{l}{} & \multicolumn{1}{l}{} & \multicolumn{1}{l}{} & \multicolumn{1}{l}{} & \multicolumn{1}{l}{} & \multicolumn{1}{l}{} & \multicolumn{1}{l}{} & \multicolumn{1}{l}{}
\end{tabular}
\end{table}

\begin{figure*}
\begin{center}
\includegraphics[width=0.8\linewidth]{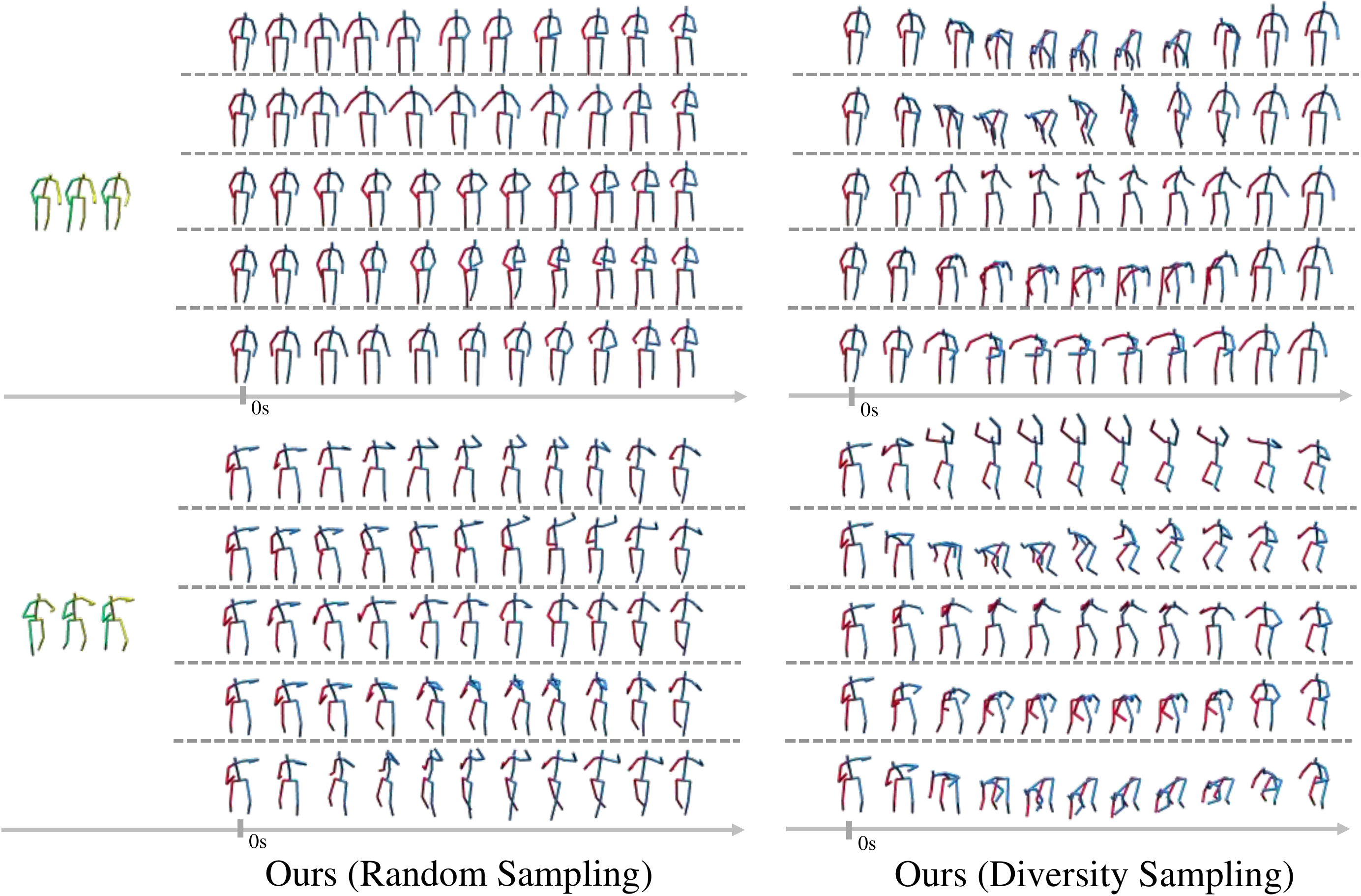}
\end{center}
\caption{Qualitative results of end posed controlled human motion prediction, by random sampling $\mathbf{z}_t$ (left) or diversity sampling $\mathbf{z}_t$ (right) on two test sequences. Each row displays a motion sample with different frames. }
\label{fig:sup_fig_3}
\end{figure*}

\noindent \textbf{Diverse Prediction.} We now focus on evaluating the controlling ability considering sample diversity with respect to the generated partial-body motions to mainly compare with \cite{yuan2020dlow,mao2021generating}, which are the closet state-of-the-art methods to ours. To be consistent with \cite{yuan2020dlow,mao2021generating}, we generate 10 samples on each test sequence (i.e., $K=10$), then promote upper-body motions to be diverse and simultaneously aim to strictly control the lower bodies (SC-L). Therefore, we set the BPI as the lower body $\mathbf{x}_l$ to train our model and then generate motion predictions with \textit{diversity sampling} $\mathbf{z}_t$ via our sampler.
$\mathbf{z}_b$ is kept still to fix the lower-body motions. Furthermore, we introduce an additional evaluation metric: Minimum Pairwise Distance (MPD), as a similarity indicator: ${\min}_{i \neq j \in \lbrace1,\dots,K\rbrace}\norm{\mathbf{x}^{i}-\mathbf{x}^{j}}_2$. MPD measures the similarity for the closest two sequences among all the generation pairs, and a lower MPD indicates higher sample resemblance.  
We can observe in Tab. \ref{tab:tab2} that our method outperforms the state-of-the-arts \cite{yuan2020dlow,mao2021generating} in the upper-body APD, with a fairly low APD for the lower body. Specifically, although Dlow \cite{yuan2020dlow} also performs diversity sampling, the hard reconstruction constraint that forces lower-body motions to be close can complicate the optimization in terms of the diversity objective.
Additionally, it is worth mentioning that compared with the remarkable APD increase for the upper body of our model (Tab. \ref{tab:tab1}: 5.782, to Tab. \ref{tab:tab2}: 15.426), the change in the lower-body APD (0.392 to 0.736) is minor, which again, demonstrates the disentangling capacity on the strictly controlling strength of $\mathbf{z}_b$. For the MPD metric, it can be seen in Tab. \ref{tab:tab2} that our method consistently outperforms the state of the arts by a large margin, which proves that our predictions yield diversity within both overall generation and arbitrary sample pairs. In particular, the diversified predictions in Gsps \cite{mao2021generating} and Dlow \cite{yuan2020dlow} suffer from severe sample duplication. This is because the network escapes to trivial solutions to optimize the mean-based diversity objective \cite{ma2021likelihood}. The qualitative comparison in Fig. \ref{fig:res3}(magenta and orange boxes) further visually supports this. Despite finishing similarly, the predictions via our sampler (purple and green, or blue and brown boxes in Fig. \ref{fig:res3}) exhibit a wide range of intermediate motions. We provide more results in the APPENDIX D.

\begin{table*}[t]
\centering
\caption
{Evaluation of the performance of end pose controlled human motion prediction via APD and MPD. The results via our method are generated with fixed $\mathbf{z}_b$ and then randomly (third column) or diversity (final column) sampling $\mathbf{z}_t$.}
\label{tab:tab4}
\begin{tabular}{ccccc}
\toprule
                     & CVAE                 & \begin{tabular}[c]{@{}c@{}}Ours\\ random sampling\end{tabular} & Dlow \cite{yuan2020dlow}                & \begin{tabular}[c]{@{}c@{}}Ours\\ diversity sampling \end{tabular} \\ \hline
APD (End pose) $\downarrow$      & 0.893                & \textbf{0.164}                                             & 0.607                & 0.471                                                     \\
APD (Whole seq.) $\uparrow$ & 6.631                & 4.666                                                      & 12.739               & \textbf{14.675}                                           \\
MPD (Whole seq.) $\uparrow$         & -                    & -                                                          & 4.548                & \textbf{11.472}                                           \\ \bottomrule
\multicolumn{1}{l}{} & \multicolumn{1}{l}{} & \multicolumn{1}{l}{}                                       & \multicolumn{1}{l}{} & \multicolumn{1}{l}{}                                      \\
\multicolumn{1}{l}{} & \multicolumn{1}{l}{} & \multicolumn{1}{l}{}                                       & \multicolumn{1}{l}{} & \multicolumn{1}{l}{}                                      \\
\multicolumn{1}{l}{} & \multicolumn{1}{l}{} & \multicolumn{1}{l}{}                                       & \multicolumn{1}{l}{} & \multicolumn{1}{l}{}                                     
\end{tabular}
\end{table*}

\begin{table}[t]
\centering
\caption
{Evaluation of the sequence length $l$ of $\mathbf{x}_{aux}$ in training our model on the controlling performance via APD. The original sequence $\mathbf{x}$ is with the length of 100 frames. Each $\mathbf{x}_{aux}$ with sequence length $l$ consists of the same start and end pose as the original sequence $\mathbf{x}$, with the intermediate $l-2$ frames completed with linear interpolation.  All the results are generated via randomly sampling $\mathbf{z}_t$ with fixed $\mathbf{z}_b$.  }
\label{tab:tab5}
\begin{tabular}{cccccc}
\toprule
                     & $l$ = 50               & $l$ = 75               & $l$ = 100              & $l$ = 125              & $l$ = 150              \\ \hline
APD (End pose) $\downarrow$      & 0.129                & 0.116                & 0.164                & 0.107                & \textbf{0.101}                \\
APD (Whole seq.) $\uparrow$ & 3.835                & 3.843                & \textbf{4.666}                & 3.751                & 3.762                \\ \bottomrule
\multicolumn{1}{r}{} & \multicolumn{1}{l}{} & \multicolumn{1}{l}{} & \multicolumn{1}{l}{} & \multicolumn{1}{l}{} & \multicolumn{1}{l}{}
\end{tabular}
\end{table}

\noindent	\textbf{Ablation Study on Sample Quality.} To verify the realism of the diverse predictions generated by our sampler, we ablate or adopt the normalizing flow-based validity objective $L_{vid}$ to compare its impact on the sample quality. As suggested in \cite{mao2021generating}, we directly calculates the negative-log-likelihood (NLL) of generated poses from pre-trained flow network $f^{*}$. It can be verified from Tab. \ref{tab:tab3} that training our sampler with $L_{vil}$ acts as a positive role in achieving the lowest NLL score, though slightly sacrificing some diversity. 

\subsection{Evaluation of End Pose Controlled Prediction}
Our model realizes another form of control by inputting the bottom path with an interpolated sequence $\mathbf{x}_{aux}$ to empower $\mathbf{z}_b$ to control the motion end pose. As such, $\mathbf{z}_t$ then governs the intermediate motions to reach such a finish pose. Note that the network architecture for this form of control remains the same as the upper/lower body control case.

Furthermore, we can diversify the intermediate motions by using our sampler for diversity sampling $\mathbf{z}_t$, but controlling all of the generated motions to end similarly via fixing $\mathbf{z}_b$. Since Dlow \cite{yuan2020dlow} can also be designed to fulfill the requirement of controlling the end pose to be similar via modifying the reconstruction objective, we then re-train Dlow as a baseline to compare with ours.

We quantitatively include the controlling ability over the end pose and motion diversity across the whole sequence in Tab. \ref{tab:tab4}. We can confirm from Tab. \ref{tab:tab4}(third column) that our method is capable of controlling the end pose, whereas CVAE cannot achieve this. Also, our method outperforms Dlow in both APD and MPD metrics after applying our diversity sampler. The reason can be that the reconstruction term in the Dlow training objective serves as a hard constraint that lowers the diversity gain. The qualitative results by our model (fix $\mathbf{z}_b$, randomly sample $\mathbf{z}_t$), and by applying diversity sampler (fix $\mathbf{z}_b$, diversity sample $\mathbf{z}_t$) are shown in Fig. \ref{fig:sup_fig_3}.
The results in Fig. \ref{fig:sup_fig_3} also visually evidence the capability of our model in controlling the end pose. Also, fixing $\mathbf{z}_t$ but varying $\mathbf{z}_b$ does not show any power of control, since the end pose for each generated motion is not kept fixed.

\noindent	\textbf{Ablation Study on Sequence Length of $\mathbf{x}_{aux}$.}
We further investigate the sequence length of $\mathbf{x}_{aux}$ on the generation performance to gain more insights into its impact. Since $\mathbf{x}_{aux}$ is assumed to start and finish at the same pose with the true future motion $\mathbf{x}$, we then change the length for the intermediate linear interpolation and train our model under five length settings of $\mathbf{x}_{aux}$ to compare the corresponding APD.  As shown in Tab. \ref{tab:tab5}, despite the slightly improved APD for the end pose, training with an unequal length design of $\mathbf{x}_{aux}$ and $\mathbf{x}$ results in a large APD drop for the whole sequence, thus limiting the diversity for intermediate variations. Based on this observation, we interpolate $\mathbf{x}_{aux}$ to keep a same length with $\mathbf{x}$ for decent variation coverage.

\section{Conclusion}
In this paper, we proposed to realize controllable human motion under a novel motion disentanglement framework. Our network models realistic human motion, and provides control for desired body parts in one training. We also extended a sampling method and adapted it to our pre-trained decoder to diversify the predictions. Moreover, we showed that a flexible design for our bottom CVAE path input further contributes to other forms of control in human motion prediction, such as controlling the end pose. Experiments qualitatively and quantitatively demonstrated that our method achieves state-of-the-art performance in predicting controllable human motion. As a limitation, our method currently resorts to an additional sampler for diversity. We will focus on modeling motion controllability, validity, and diversity under a unified framework in the future.

\appendices
\section{Network Architecture}

\begin{figure}
\begin{center}
\includegraphics[width=1.0\linewidth]{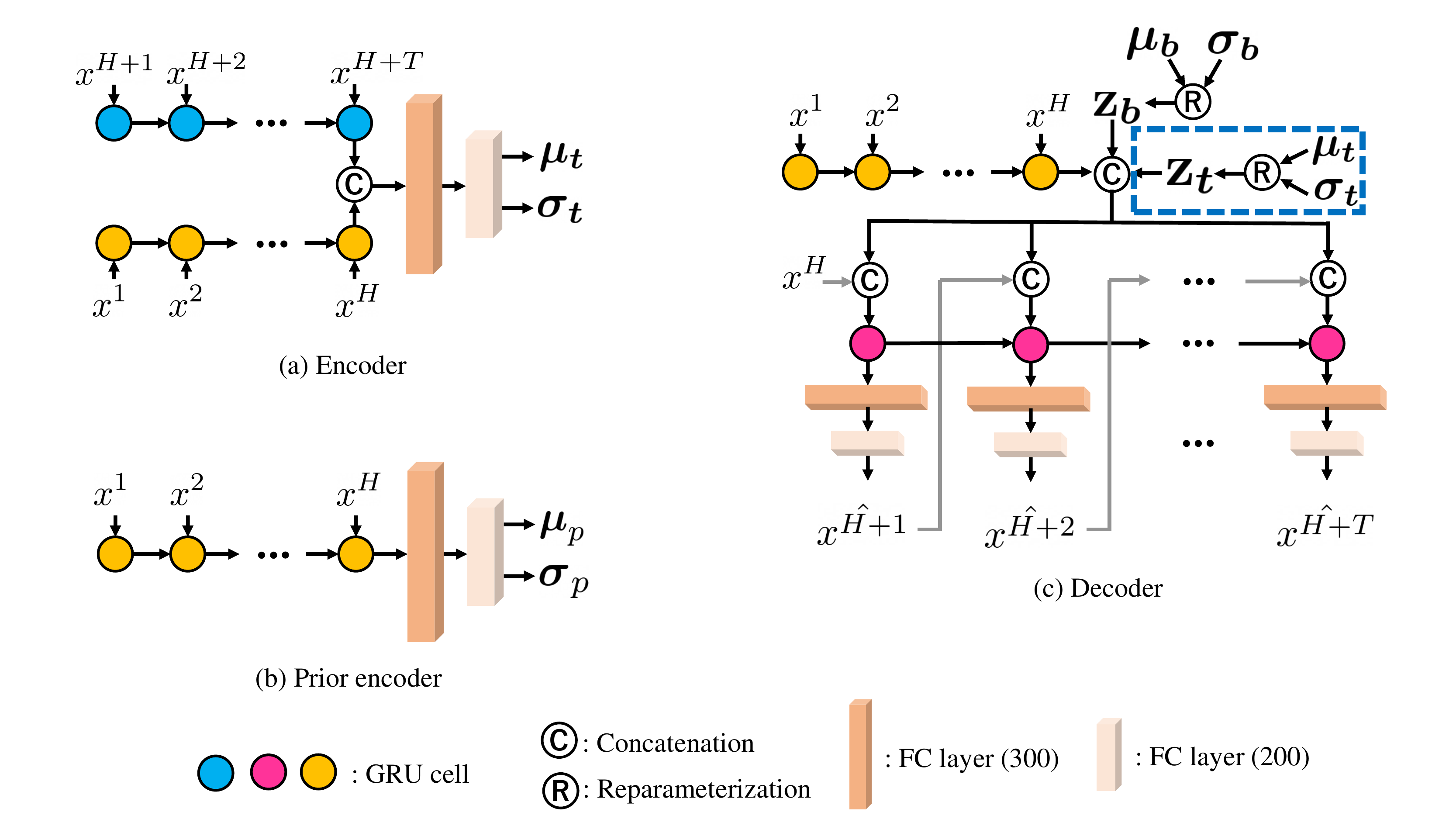}
\end{center}
\caption{Detailed architectures for each sub-networks of our model.}
\label{fig:sup_fig_1}
\end{figure}

Both the top path and the bottom path of the proposed network architecture are composed of an encoder, a prior encoder, and a decoder. Fig. \ref{fig:sup_fig_1} shows the detailed structure of (prior) encoder/decoder used in the top path. The bottom path uses the exact same construction as the top path except that the decoder of the bottom path does not include the concatenation operation in the blue dashed area in Fig. \ref{fig:sup_fig_1}. Besides, $\boldsymbol{\mu_b}$ and $\boldsymbol{\sigma_b}$ in subplot (c) of Fig. \ref{fig:sup_fig_1} are the Gaussian parameters regressed by the bottom path encoder. FC layer ($n$) refers to a fully-connected layer with output size of $n$. All the sizes of Gated Recurrent Unit (GRU) cell, Gaussian parameters, and latent variables are equally set to 128.

\section{Derivation of the Training Objective }
We here show the detailed derivation of our optimization objective (i.e., Eq. \ref{eq:eq7}). The optimization target for our model is to \textit{maximize} the likelihood of the given data. Following the special generative process for full-body motion $\mathbf{x}$ and the motion for only one portion of the human body $\mathbf{x}^{1}_{p}$, in which $\mathbf{x}$ and $\mathbf{x}^{1}_{p}$ are assumed to be independently generated, we can define the log-likelihood (LL) of $p(\mathbf{x},\mathbf{x}^{1}_{p} | \mathbf{c}, \mathbf{c}^{1}_{p})$ as:

\begin{equation}
\small
\label{eq:eq21}
\begin{aligned}
LL &= \log p(\mathbf{x},\mathbf{x}^{1}_{p} | \mathbf{c}, \mathbf{c}^{1}_{p}) \\ 
&= \log p(\mathbf{x}| \mathbf{c})p(\mathbf{x}^{1}_{p} | \mathbf{c}^{1}_{p})\\ 
&= \log \int_{\mathbf{z}_t}\int_{\mathbf{z}_b} p(\mathbf{x}|\mathbf{z}_t,\mathbf{z}_b, \mathbf{c}) p(\mathbf{x}^{1}_{p}|\mathbf{z}_b, \mathbf{c}^{1}_{p})p(\mathbf{z}_t|\mathbf{c})p(\mathbf{z}_b|\mathbf{c}^{1}_{p})d\mathbf{z}_td\mathbf{z}_b \\
&= \log \int_{\mathbf{z}_t}\int_{\mathbf{z}_b} {q(\mathbf{z}_t|\mathbf{x},\mathbf{c})q(\mathbf{z}_b|\mathbf{x}^{1}_{p},\mathbf{c}^{1}_{p})} \\ 
& \quad \times \frac{p(\mathbf{x}|\mathbf{z}_t,\mathbf{z}_b, \mathbf{c}) p(\mathbf{x}^{1}_{p}|\mathbf{z}_b, \mathbf{c}^{1}_{p})p(\mathbf{z}_t|\mathbf{c})p(\mathbf{z}_b|\mathbf{c}^{1}_{p})}{q(\mathbf{z}_t|\mathbf{x},\mathbf{c})q(\mathbf{z}_b|\mathbf{x}^{1}_{p},\mathbf{c}^{1}_{p})}d\mathbf{z}_td\mathbf{z}_b.\\
\end{aligned}
\end{equation}
By using Jensen's inequality, we follow the VAE optimization strategy of employing the Evidence Lower BOund (ELBO) as the surrogate to enable a tractable learning:
\begin{equation}
\small
\label{eq:eq22}
\begin{aligned}
&LL \geq \int_{\mathbf{z}_t}\int_{\mathbf{z}_b}  {q(\mathbf{z}_t|\mathbf{x},\mathbf{c})q(\mathbf{z}_b|\mathbf{x}^{1}_{p},\mathbf{c}^{1}_{p})} 
\\ 
& \quad \times \log  \frac{p(\mathbf{x}|\mathbf{z}_t,\mathbf{z}_b, \mathbf{c}) p(\mathbf{x}^{1}_{p}|\mathbf{z}_b, \mathbf{c}^{1}_{p})p(\mathbf{z}_t|\mathbf{c})p(\mathbf{z}_b|\mathbf{c}^{1}_{p})}{q(\mathbf{z}_t|\mathbf{x},\mathbf{c})q(\mathbf{z}_b|\mathbf{x}^{1}_{p},\mathbf{c}^{1}_{p})}d\mathbf{z}_td\mathbf{z}_b \\
&= \mathbb{E}_{q(\mathbf{z}_t|\mathbf{x},\mathbf{c}),q(\mathbf{z}_b|\mathbf{x}^{1}_{p},\mathbf{c}^{1}_{p})}[\log \frac{p(\mathbf{x}|\mathbf{z}_t,\mathbf{z}_b, \mathbf{c}) p(\mathbf{x}^{1}_{p}|\mathbf{z}_b, \mathbf{c}^{1}_{p})p(\mathbf{z}_t|\mathbf{c})p(\mathbf{z}_b|\mathbf{c}^{1}_{p})}{q(\mathbf{z}_t|\mathbf{x},\mathbf{c})q(\mathbf{z}_b|\mathbf{x}^{1}_{p},\mathbf{c}^{1}_{p})}] \\
&= \mathbb{E}_{q(\mathbf{z}_t|\mathbf{x},\mathbf{c}),q(\mathbf{z}_b|\mathbf{x}^{1}_{p},\mathbf{c}^{1}_{p})}[\log p(\mathbf{x}|\mathbf{z}_t,\mathbf{z}_b, \mathbf{c}) + \log p(\mathbf{x}^{1}_{p}|\mathbf{z}_b, \mathbf{c}^{1}_{p}) \\
& \quad + \log \frac{p(\mathbf{z}_t|\mathbf{c})}{q(\mathbf{z}_t|\mathbf{x},\mathbf{c})} + \log \frac{p(\mathbf{z}_b|\mathbf{c}^{1}_{p})}{q(\mathbf{z}_b|\mathbf{x}^{1}_{p},\mathbf{c}^{1}_{p})}  ] \\
&= \mathbb{E}_{q(\mathbf{z}_t|\mathbf{x},\mathbf{c}),q(\mathbf{z}_b|\mathbf{x}^{1}_{p},\mathbf{c}^{1}_{p})}[\log p(\mathbf{x}|\mathbf{z}_t,\mathbf{z}_b, \mathbf{c})] \\ 
& \quad + \mathbb{E}_{q(\mathbf{z}_b|\mathbf{x}^{1}_{p},\mathbf{c}^{1}_{p})}[\log p(\mathbf{x}^{1}_{p}|\mathbf{z}_b, \mathbf{c}^{1}_{p})] \\
& \quad+ \mathbb{E}_{q(\mathbf{z}_t|\mathbf{x},\mathbf{c})}[\log \frac{p(\mathbf{x}|\mathbf{z}_t)}{q(\mathbf{z}_t|\mathbf{x},\mathbf{c})}] + \mathbb{E}_{q(\mathbf{z}_b|\mathbf{x}^{1}_{p},\mathbf{c}^{1}_{p})}[\log \frac{p(\mathbf{x}^{1}_{p}|\mathbf{z}_b)}{q(\mathbf{z}_b|\mathbf{x}^{1}_{p},\mathbf{c}^{1}_{p})}].
\end{aligned}
\end{equation}

\noindent	By including the network parameters, we then obtain the final training objective as 
\begin{equation}
\label{eq:eq23}
\begin{aligned}
L = &\mathbb{E}_{q_{\phi}(\mathbf{z}_t|\mathbf{x},\mathbf{c}), q_{{\phi}^{\prime}}(\mathbf{z}_b|\mathbf{x}^{1}_{p},\mathbf{c}^{1}_{p})}[\log p_{\theta}(\mathbf{x}|\mathbf{z}_t, \mathbf{z}_b, \mathbf{c})] \\ 
&+ \mathbb{E}_{q_{{\phi}^{\prime}}(\mathbf{z}_b|\mathbf{x}^{1}_{p},\mathbf{c}^{1}_{p})}[\log p_{{\theta}^{\prime}}(\mathbf{x}^{1}_{p}|\mathbf{z}_b, \mathbf{c}^{1}_{p})] \\
&-  \mathrm{KL}(q_{\phi}(\mathbf{z}_t|\mathbf{x},\mathbf{c})||p_{\psi}(\mathbf{z}_t|\mathbf{c})) \\ 
&- \mathrm{KL}(q_{{\phi}^{\prime}}(\mathbf{z}_b|\mathbf{x}^{1}_{p},\mathbf{c}^{1}_{p})||p_{\hat{\psi}}(\mathbf{z}_b|\mathbf{c}^{1}_{p})).
\end{aligned}
\end{equation}

\begin{figure*}
\begin{center}
\includegraphics[width=0.8\linewidth]{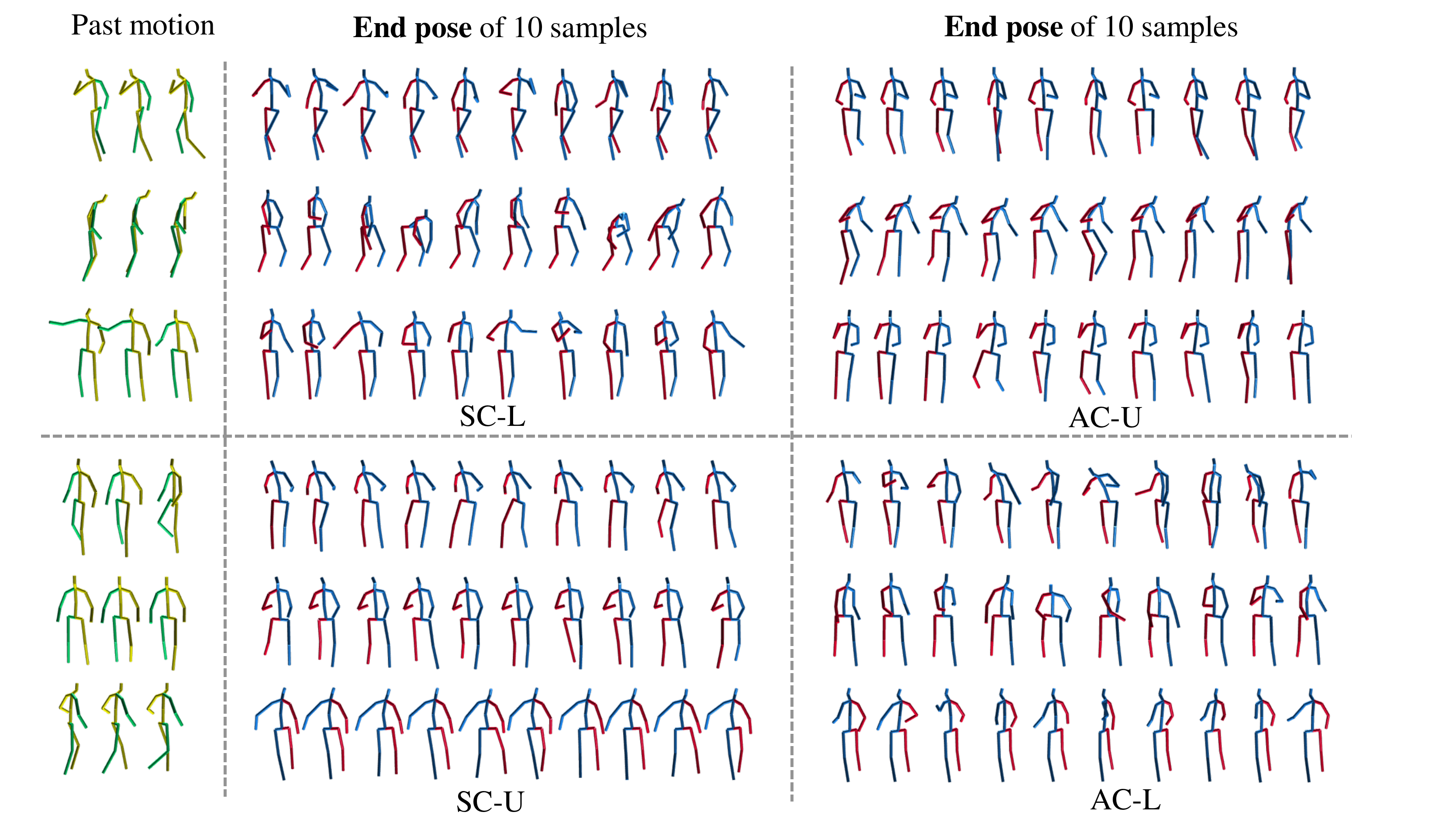}
\end{center}
\caption{Additional results for strictly controlled (SC) or adaptively controlled (AC) predictions on lower (L)/upper (U) partial body. We show three past frames (left), and the end pose of 10 samples (right) via our trained top path decoder by randomly sampling the latent variable pair ($\mathbf{z}_t, \mathbf{z}_b$). }
\label{fig:sup_fig_4}
\end{figure*}

\begin{figure*}
\begin{center}
\includegraphics[width=0.8\linewidth]{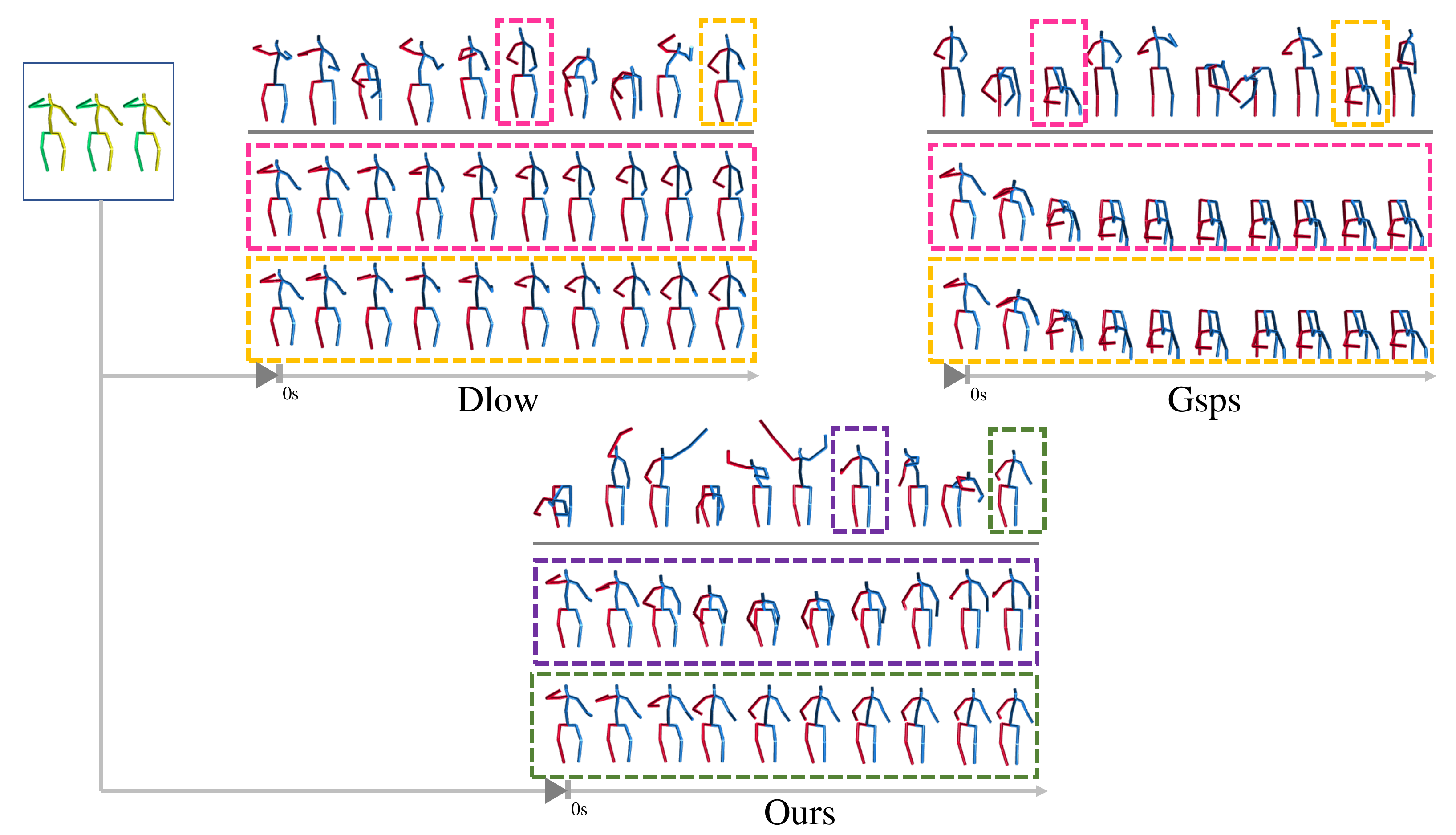}
\includegraphics[width=0.8\linewidth]{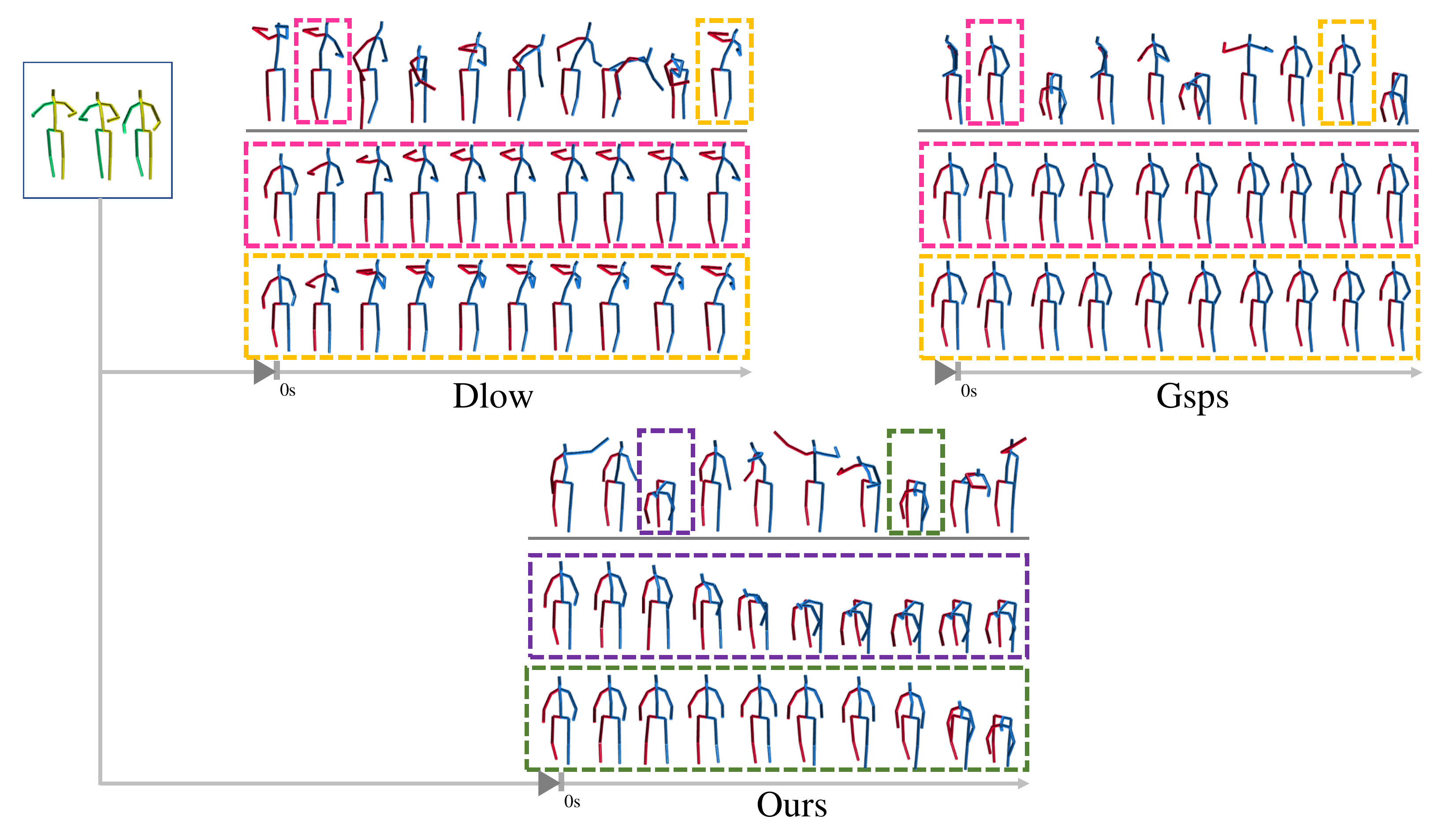}
\end{center}
\caption{Additional results for diverse motion prediction. All the methods aim to predict strictly controlled lower-body (SC-L) motions with diverse upper-body motions. Our results is obtained by fixing $\mathbf{z}_b$, then diversity sampling $\mathbf{z}_t$ via our sampler. The first row for each method exhibits the end pose of 10 samples. Below we visualize two example sequences that end similarly.   }
\label{fig:sup_fig_5}
\end{figure*}

\section{Validity Objective for the Diversity Sampler}
We here detail the validity objective for our sampler. We adopt the normalizing flow-based pose prior in \cite{mao2021generating} to ensure sampling valid poses. The design motivation of a pose prior is based on the observation that human motion is composed of smooth poses, and it is much easier to obtain a sufficient amount of poses than motions to learn the pose-level kinematics for human. Normalizing flow \cite{rezende2015variational} is a type of generative model which has gained growing attention recently in human motion modeling \cite{henter2020moglow,huang2018neural}. Precisely, normalizing flow learns a bijective transformation $f(\cdot)$ between an unknown 3D human pose distribution $p(x)$ and a tractable density function (e.g., Gaussian), such that the learned mapping function bridges a sampled  pose from the complex distribution $x \sim p(x)$ and the base Gaussian function $\bm{o} \sim \mathcal{N}(0,I)$ in the manner of $\bm{o} = f(x)$. The mapping function $f(\cdot)$ is parameterized with deep neural networks, and can be trained by minimizing the negative log-likelihood (NLL) of the samples:
\begin{equation}
\label{eq:eq24}
\begin{aligned}
NLL &= -\underset{x \in \mathcal{S}}{\sum}\log p(x) \\
& = -\underset{x \in \mathcal{S}}{\sum}\log g(\bm{o}) - \log \left| \det(\frac{\partial f}{\partial x}) \right|,
\end{aligned}
\end{equation}
where $g(\bm{o}) = \mathcal{N}(\bm{o}|0,I)$, and the $\left| \det(\frac{\partial f}{\partial x}) \right|$ indicates the absolute-determinant of the Jacobian matrix of $f(\cdot)$. $\mathcal{S}$ is the dataset for all the valid human poses. The trained network $f^{*}$ can thus be leveraged to motivate the sampler to sample realistic poses. We define the validity objective $L_{vli}$ as the NLL for a generated human pose $\hat{x}$ as:
\begin{equation}
\label{eq:eq25}
L_{vli} = -\log g(\hat{\bm{o}}) - \log \left| \det(\frac{\partial f^{*}}{\partial \hat{x}}) \right|,
\end{equation}
in which $\hat{\bm{o}} = f^{*}(\hat{x})$ and $g(\hat{\bm{o}}) =  \mathcal{N}(\hat{\bm{o}}|0,I)$.

Based on the observation that a valid human pose relies largely on the kinematics of joint angles, the distribution of limb directions is modeled in learning the pose prior, instead of directly modeling the 3D joint coordinates for human poses. In particular, the limb direction $l_i \in \mathbb{R}^{3}$ for the $i$-th joint can be calculated as:
\begin{equation}
\label{eq:eq26}
\begin{aligned}
l_i = \frac{j_i - j^p_i}{\norm{j_i - j^p_i}_2},
\end{aligned}
\end{equation}
where $j_i \in \mathbb{R}^{3}$ is the coordinate for the $i$-th joint, and $j^p_i \in \mathbb{R}^{3}$ represents the corresponding parent joint coordinate. We then redefine the human pose $x$ with the set of all human limb directions $x = [l^{T}_1,\dots, l^{T}_N]$ as a substitution for the joint locations. $N$ denotes the joint number, which is equal to the number for limb directions.

The network to model $f$ is simply designed based on fully-connected (FC) layers. In implementation, the QR decomposition and a monotonic activation function are leveraged to ensure the network is invertible. Specifically, each FC layer is designed as
\begin{equation}
\label{eq:eq27}
\begin{aligned}
U = \mathrm{P}(V\boldsymbol{W} + b),
\end{aligned}
\end{equation}
in which $U \in \mathbb{R}^{d_j}$ and $V \in \mathbb{R}^{d_j}$ are the output and input feature vectors with the same number of dimension $d_j = 3N$, respectively. Following QR decomposition, the weights $\boldsymbol{W}$ of each layer is further decomposed via $\boldsymbol{W=QR}$, where $\boldsymbol{Q} \in \mathbb{R}^{d_j \times d_j}$ is an orthogonal matrix, and $\boldsymbol{R} \in \mathbb{R}^{d_j \times d_j}$ is an upper triangular matrix whose diagonal elements are positive. $b \in \mathbb{R}^{d_j}$ represents the bias, and $\mathrm{P}$ is the PReLU activation. The FC layer is repeated for three times to deepen the structure.

\section{More Qualitative Results}
We provide more qualitative results of our method (random sampling) in Fig. \ref{fig:sup_fig_4}. More comparisons with the state-of-the-arts are shown in Fig. \ref{fig:sup_fig_5}, in which our results are obtained by diversity sampling via our sampler.

\ifCLASSOPTIONcaptionsoff
  \newpage
\fi



%
\bibliographystyle{abbrv}
\bibliography{main}

%






\end{document}